\documentclass[letterpaper]{article} % DO NOT CHANGE THIS
\usepackage{aaai24}  % DO NOT CHANGE THIS
\usepackage{times}  % DO NOT CHANGE THIS
\usepackage{helvet}  % DO NOT CHANGE THIS
\usepackage{courier}  % DO NOT CHANGE THIS
\usepackage[hyphens]{url}  % DO NOT CHANGE THIS
\usepackage{graphicx} % DO NOT CHANGE THIS
\usepackage{natbib}  % DO NOT CHANGE THIS AND DO NOT ADD ANY OPTIONS TO IT
\usepackage{caption} % DO NOT CHANGE THIS AND DO NOT ADD ANY OPTIONS TO IT
\usepackage{algorithm2e}
\usepackage{amstext}
\usepackage{amsmath}
\usepackage{multirow}
\usepackage{xcolor}
\usepackage{etoolbox,lipsum}
\newcommand{\Besizer}{B$\mathrm{\acute{e}}$zier }
\newcommand{\besizer}{b$\mathrm{\acute{e}}$zier }
\DeclareRobustCommand\onedot{\futurelet\@let@token\@onedot}
\def\etal{\emph{et al}.}

\usepackage{newfloat}
\usepackage{listings}
\DeclareCaptionStyle{ruled}{labelfont=normalfont,labelsep=colon,strut=off} % DO NOT CHANGE THIS
\title{Optimize and Reduce: A Top-Down Approach for Image Vectorization}
\author{
    Or Hirschorn\equalcontrib,
    Amir Jevnisek\equalcontrib,
    Shai Avidan
}
\affiliations{
    School of Electrical Engineering\\
    Tel-Aviv University \\
    Tel-Aviv, Israel \\
    orhirschorn@mail.tau.ac.il, amirjevn@mail.tau.ac.il, avidan@eng.tau.ac.il
}

\begin{document}

\maketitle
\begin{abstract}
Vector image representation is a popular choice when editability and flexibility in resolution are desired. However, most images are only available in raster form, making raster-to-vector image conversion (vectorization) an important task. 
Classical methods for vectorization are either domain-specific or yield an abundance of shapes which limits editability and interpretability.
Learning-based methods, that use differentiable rendering, have revolutionized vectorization, at the cost of poor generalization to out-of-training distribution domains, and optimization-based counterparts are either slow~\cite{ma2022towards} or produce non-editable and redundant shapes~\cite{li2020differentiable}. 
In this work, we propose Optimize \& Reduce (O\&R), a top-down approach to vectorization that is both fast and domain-agnostic. O\&R aims to attain a \emph{compact} representation of input images by iteratively optimizing \Besizer curve parameters and significantly reducing the number of shapes, using a devised importance measure. 
We contribute a benchmark of five datasets comprising images from a broad spectrum of image complexities - from emojis to natural-like images. Through extensive experiments on hundreds of images, we demonstrate that our method is domain agnostic and outperforms existing works in both reconstruction and perceptual quality for a fixed number of shapes. Moreover, we show that our algorithm is $\times 10$ faster than the state-of-the-art optimization-based method\footnote{Code: \url{https://github.com/ajevnisek/optimize-and-reduce}}.
\end{abstract}

\section{Introduction}
\begin{figure}[t]
\centering
\begin{tabular}{c}
    MSE vs Number of Vector Shapes \\
    \includegraphics[width=0.45\textwidth]{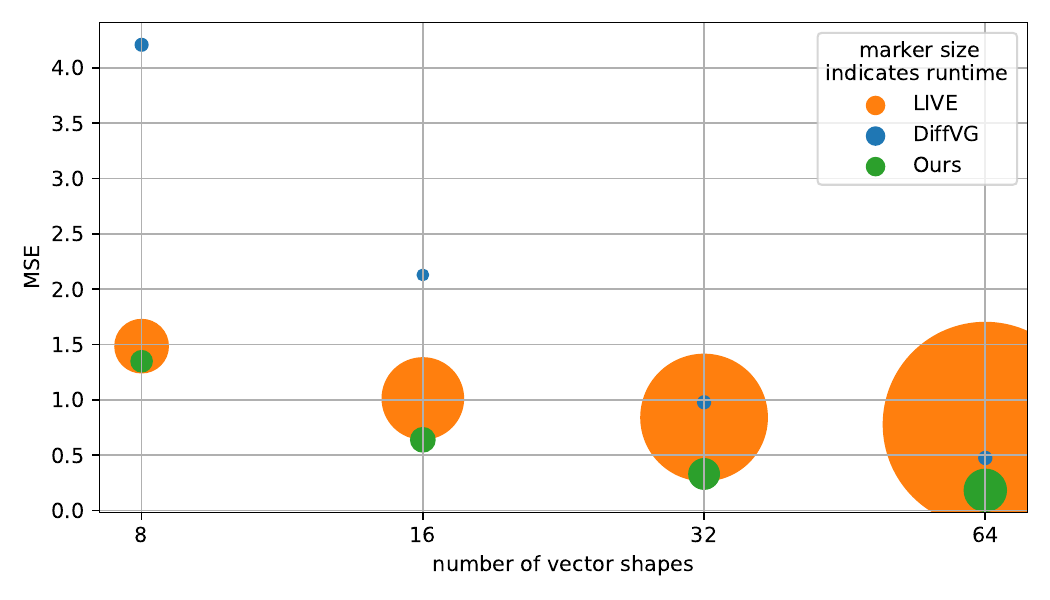} \\
    - Runtime indicated by marker-size ($\downarrow$ better)
     
\end{tabular}
\caption{\textbf{Optimize \& Reduce}: A fast Top-Down approach for raster to vector image conversion, aiming at a low budget of shapes for vectorization. Our method (green circles) achieves lowest MSE error ($y$-axis) for all shape counts ($x$-axis) with a low runtime cost (size of circle).
}

% \caption{\textbf{Optimize \& Reduce}: A fast Top-Down approach for raster to vector image conversion, aiming at a low shapes budget vectorization. Our method achieves a good tradeoff between runtime performance and reconstruction quality on the Emoji benchmark. As we are comparable to LIVE in terms of reconstruction loss, we shift the operation curve towards DiffVG in runtime, making the use of our vectorization algorithm much more practical. 
% }

\label{fig:live_with_runtime_after_refactor}
\end{figure}

\begin{figure}
  \centering
  \begin{tabular}{cccc}
    \includegraphics[width=0.09\textwidth]{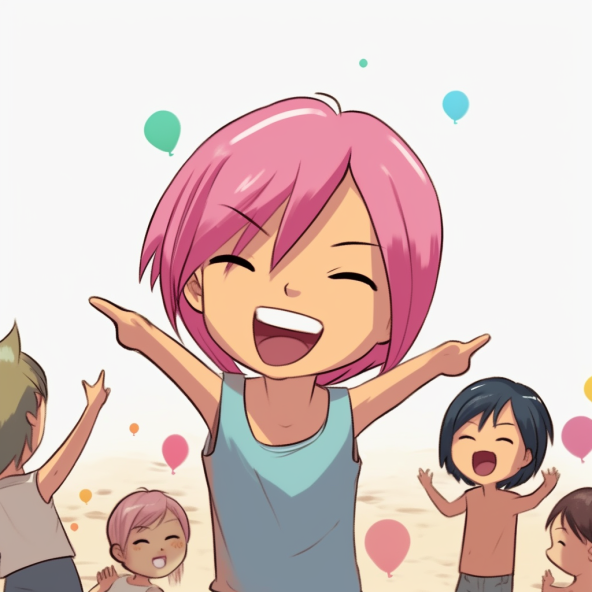} & 
    \includegraphics[width=0.09\textwidth]{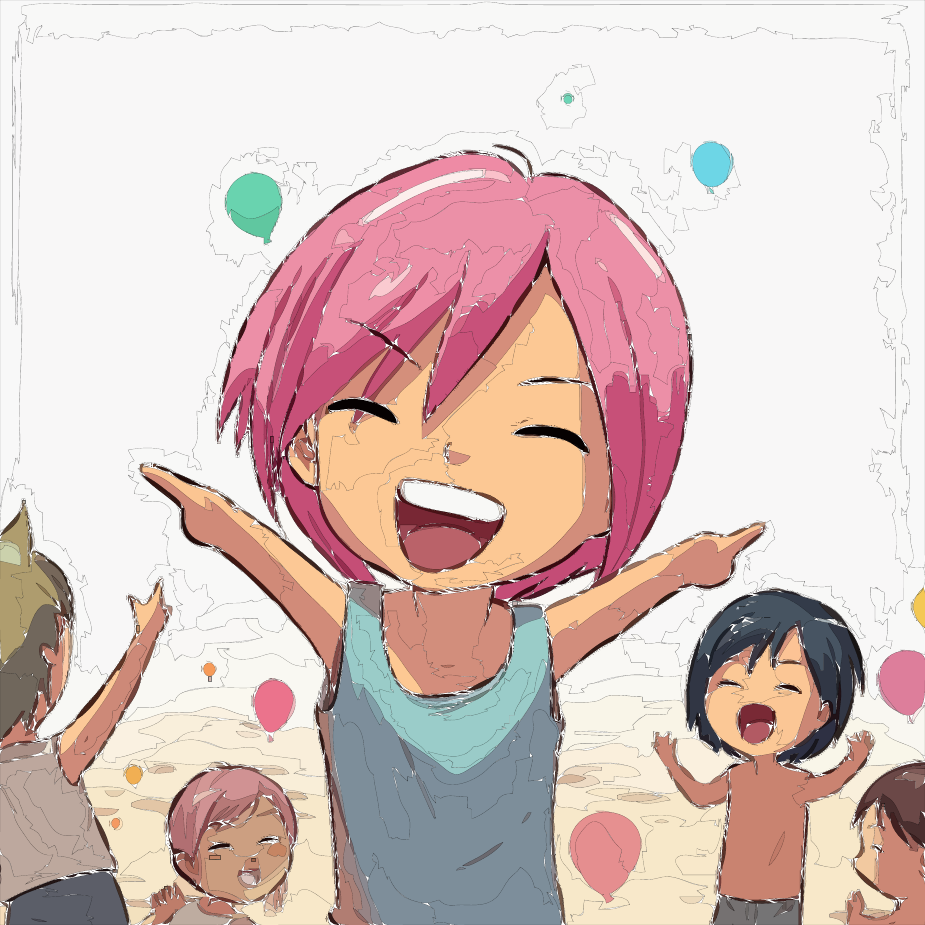} &
    \includegraphics[width=0.09\textwidth]{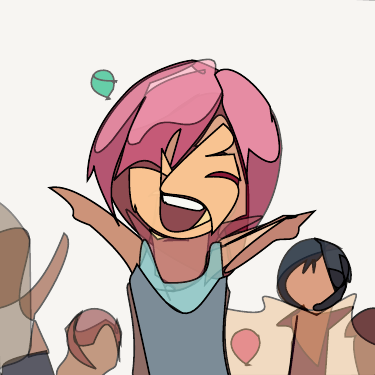} &
    \includegraphics[width=0.09\textwidth]{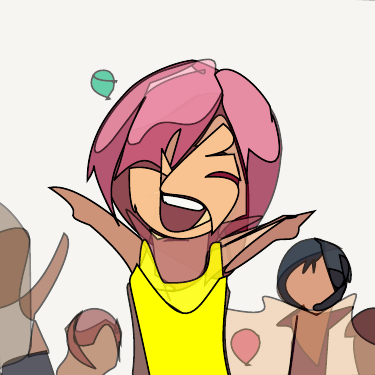} \\
    (a) & (b) & (c) & (d) \\
    \multirow{3}{*}{Raster Image} & AutoTracer  & \textbf{Ours} & \textbf{Ours} \\
      &  \multirow{2}{*}{4579 shapes} & \multirow{2}{*}{32 shapes} & 2 shapes  \\
      & & & color edit \\

    \end{tabular}
  \caption{\textbf{Low Shapes Budget Vectorization:} Low budget shapes makes editing easier.
  }
  \label{fig:motivation}
\end{figure}
\begin{figure*}
\centering
\includegraphics[width=0.98\textwidth]{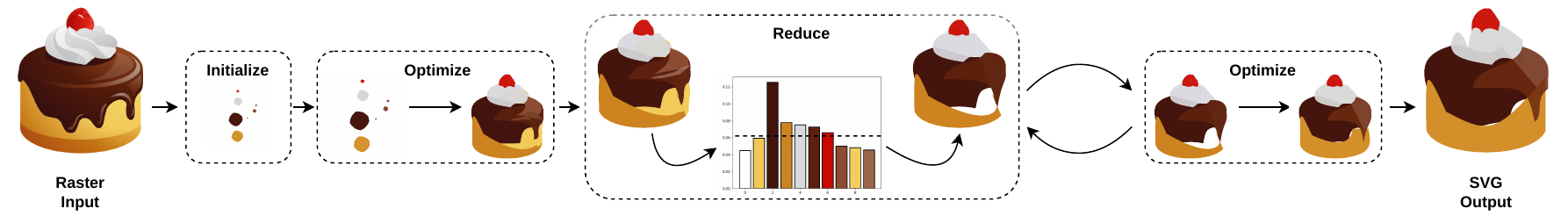}
\caption{\textbf{Optimize \& Reduce:} 
We start with a raster image and use DBSCAN~\cite{ester1996density} and connected components to initialize a large number of \Besizer shapes. We \textbf{Optimize} the shapes using DiffVG~\cite{li2020differentiable}. Then we \textbf{Reduce} the number of shapes by ranking them according to $\mathcal{L}_{reduce}$. That is, the least important $k$ shapes (shapes with scores below the dashed line in the bar graph) are removed. The remaining shapes serve as an initial guess for the next optimization iteration.}
\label{fig:scheme}
\end{figure*}

Vectorization refers to the process of transforming a raster image into a vector image. While it is possible to create a vector image with a shape for each pixel, ensuring a perfect reconstruction, this approach lacks compactness and editability. The question arises: can we achieve compact vectorization without compromising quality?

This paper aims to achieve vectorization while minimizing shape count. We show in Figure~\ref{fig:live_with_runtime_after_refactor} the tradeoff between shape count and Mean Squared Error (MSE) for our method and two comparative techniques. Our approach demonstrates better performance compared to the state-of-the-art LIVE method~\cite{ma2022towards} in both MSE and runtime aspects. Furthermore, our method surpasses DiffVG~\cite{li2020differentiable} in terms of minimizing MSE.

Vector images are an appealing image representation for those seeking infinite resolution and editability. Vector Images are made up of lines and shapes with clear mathematical formulations. A graphical renderer processes the mathematical formulations and renders the image at any specified resolution. This resolution-free property makes them suitable for cases where the image is presented in multiple resolutions, such as commercial logos. A single vector image can be resized to fit the resolution of a business card, a header image in a %company's
formal letter, a T-shirt, or a billboard.

The task of vectorization has been studied extensively for many decades~\cite{jimenez1982some}. We identify two primary drawbacks of classical algorithms for vectorization. First, they are often tailored to a specific domain of images such as fonts~\cite{itoh1993curve}, line drawings~\cite{chiang1998new, noris2013topology}, cartoons~\cite{zhang2009vectorizing} or cliparts~\cite{favreau2017photo2clipart, dominici2020polyfit}. Second, they yield an abundance of shapes making the representation complex. For instance, Potrace~\cite{selinger2003potrace} generates thousands of shapes for natural-like images. 

In contrast, our objective is to achieve a compact vector representation, characterized by a relatively small number of shapes. Intuitively, a low number of shapes enhances the editability of vector images. To demonstrate this we show a comparison between the vectorization of the classical AutoTrace~\cite{weber2004autotrace} and our vectorization approach in Figure~\ref{fig:motivation}. Performing a color edit to the shirt is straightforward when the number of shapes is low (32 shapes), whereas, with AutoTrace the task becomes challenging due to the need to augment hundreds of shapes.

Differentiable rasterizers, such as DiffVG~\cite{li2020differentiable}, marked a turning point in the field. DiffVG can optimize the vector shape parameters with respect to any raster-based loss function. DiffVG suggested a vanilla vectorization method, directly optimizing the shape parameters. This was extended to two types of vectorization methods: dataset-based and optimization-based approaches. Dataset-based approaches train a neural network to regress the shape parameters and train over some datasets. These methods leverage DiffVG propagation of gradients from the loss function to the shape parameters. Though these methods enjoy a fast inference time, they generalize poorly to out-of-dataset image distribution.

Our focus in this work is to extend Optimization-based methods and provide a method that excels when vectorization runtime and shape compactness are important. While DiffVG~\cite{li2020differentiable} proposes direct optimization for curve parameters, this basic approach is highly sensitive to initialization and tends to generate intricate shapes when optimized using reconstruction or perceptual loss functions. The current State-of-the-art vectorization method in terms of reconstruction error is Layer-wise Image Vectorization (LIVE)~\cite{ma2022towards}, which employs a bottom-up process to generate a comprehensive set of vectors, achieving layerwise decomposition. However, this advantage comes at the expense of increased runtime, as depicted in Figure\ref{fig:live_with_runtime_after_refactor}.

Our approach, Optimize \& Reduce (O\&R), is a top-down method. Given a raster image, we cluster pixel values to create an initial guess of the shapes. Then, in the {\em Optimize} step, we use DiffVG to optimize all shapes.
We use both reconstruction loss, to match the input raster image, as well as a geometric loss to encourage DiffVG to generate "simple" shapes that do not intersect. In the {\em Reduce} step, we remove shapes according to an importance measure that we design. The remaining shapes serve as the initial guess for the next round of optimization. The process is iterated until the target number of shapes is reached. The combination of a smart, clustering-based initialization and our O\&R scheme allows DiffVG to optimize for all shapes {\em at once} leading to excellent results.

Through extensive experimentation, we demonstrate that our approach achieves a runtime speedup of over $\times 10$ compared to LIVE, all while maintaining reconstruction accuracy. Furthermore, we establish that DiffVG produces reconstructions that are inferior to our method, whether initialized randomly (as shown in Figure~\ref{fig:live_with_runtime_after_refactor}) or with a more informed approach (as illustrated in Figure~\ref{fig:DBSCAN_contribution_emojis}).

We contribute a benchmark of five datasets covering a wide range of image complexities. Three of these datasets consist of images that are typically of interest for vectorization, such as (two versions of) emojis, NFT-apes, and SVG images converted to raster images. The last dataset we propose consists of natural-like images generated by Midjourney. Natural images have also been shown to be of interest for vectorization in previous works~\cite{li2020differentiable, ma2022towards}. 
Thus, our paper's contribution is threefold: (1) we propose a new vectorization method, (2) through extensive experiments, we demonstrate that our method produces high-quality vector images in terms of reconstruction and perceptual quality, and (3) we provide a benchmark consisting of five diverse image types.

\section{Related Work}

\paragraph{\bf Raster To Vector Conversion}
The conventional image vectorization process typically involves two main steps. Initially, pixels are clustered into regions, either through segmentation or triangulation based on edge detection. Subsequently, these regions are fitted using vector primitives. The optimization in the second step includes both discrete and continuous variables, such as the number and position of primitives. Various exploration mechanisms, like those presented in~\cite{li2020approximating, favreau2017photo2clipart, olsen2011image} have been proposed for this purpose.
Among them, Potrace\cite{selinger2003potrace} and Diebel~\etal\cite{diebel2008bayesian} utilized an empirical two-stage algorithm to regress segmented components as polygons and bezigons.

In the deep learning era, large, pre-trained models can be employed for pixel clustering, utilizing techniques like semantic segmentation. Moreover, perceptual loss based on deep neural networks can evaluate fitness between image regions and vector primitives.
One can categorize vectorization techniques into Dataset-based and Optimization-based. 

Some works focus on dataset-level learning, which simplifies the problem to one specific domain of images. 
Early works used CNNs~\cite{zheng2019strokenet} or GANs~\cite{balasubramanian2019teaching} to learn a direct conversion between the two representations. Later, Lopes~\etal\cite{lopes2019learned} and Carlier~\etal\cite{carlier2020deepsvg} framed the synthesis of vectors as a problem of predicting sequences, whereby the image is represented as a sequence of instructions for drawing, emulating the representation of vector graphics in typical formats. 
Similarly, other works formulate this task as an image-to-image translation using paired data of corresponding images and shape vectors ~\cite{yi2019apdrawinggan, li2019im2pencil, li2019photo, muhammad2018learning}, which are difficult to acquire.
To overcome this supervision problem, there is a popular research trend of integrating differentiable rendering methods~\cite{li2020differentiable, mihai2021differentiable} with deep learning models. 
Reddy~\etal\cite{reddy2021im2vec} suggested a dataset-level method, which utilized differentiable rendering~\cite{li2020differentiable}. They used a VAE to encode the input image, and using an RNN network create latent geometric features, which are decoded using a \Besizer path decoder and a differentiable rasterizer. However, this method is limited to only a specific domain and performs poorly on complex images.
Recently, Jain~\etal\cite{jain2022vectorfusion} combined a pre-trained text-to-image diffusion model and a differentiable renderer to create a text-to-vector model. Yet, the topology of the output SVG isn't intuitive, as simple shapes are sometimes represented by multiple overlapping layers.

Meanwhile, optimization-based methods were also suggested, not relying on datasets or lengthy training phases.
Vinker~\etal\cite{vinker2022clipasso, vinker2022clipascene} proposed using semantic Clip objective to optimize \Besizer curves for object/scene sketching. 
Ma~\etal\cite{ma2022towards} presented a layer-wise image vectorization method, which converts an image to vectors while maintaining its topology, forming shapes that are semantically consistent with the human perspective. They use a differentiable renderer~\cite{li2020differentiable} in a bottom-up approach, optimizing the vector image layer-by-layer. 
However, this method is extremely slow, as optimizing each layer is compute-intensive. In addition, the output greatly varies according to the initialization method of the shapes, thus becoming a strong prior which fails for complex and texture-rich images.
In contrast, our top-down approach is fast, robust to initialization, and can handle both simple and complex images while maintaining editability.

\paragraph{\bf Topology and Editability}
To ensure easy SVG editing, proper organization, and smooth shapes are crucial. Previous studies have explored image topology for both raster images and vector shapes~\cite{guo2019deep, zhu2022tcb, kopf2011depixelizing}.

Some works study topology, to characterize fonts styles~\cite{wang2021deepvecfont, reddy2021multi, shimoda2021rendering}, while others use topology as a regularizer for vectorizing sketches~\cite{bessmeltsev2019vectorization, chan2022learning, mo2021general}. 
For images, early works focused on vectorizing cliparts using various boundary-aware methods~\cite{hoshyari2018vectorization, dominici2020polyfit, riso2022pop}. Later, Favreau~\etal\cite{favreau2017photo2clipart} and Shen~\etal\cite{shen2021clipgen} decomposed bitmap pictures into gradient-filled vectors, creating visual hierarchy through segmentation and merging.
Alternative studies like diffusion curves~\cite{xie2014hierarchical, zhao2017inverse, orzan2008diffusion} and meshes~\cite{sun2007image, xia2009patch, lai2009automatic} were explored but fell short for complex scenes.

Reddy~\etal\cite{reddy2021im2vec} proposed simultaneous image vectorization and topology extraction but struggled with shape ordering and domain specificity. Later, Ma~\etal\cite{ma2022towards} introduced a bottom-up SVG construction with geometric constraints, excelling for basic clipart but struggling with complex images.
Our method does not require any pre-segmentation or model training and exhibits a satisfying ability to explore image topologies and create easily editable SVGs, even for complex images.

\paragraph{\bf Image Abstraction}
The process of abstraction is central to creating a new image.
To effectively convey desired details, artists must choose which essential visual characteristics of the subject to depict, and which to exclude (often many). Abstract images maintain semantic essence, exhibiting simplicity with fewer details and textures.

Prior works~\cite{gao2019artistic, frans2022clipdraw, tian2022modern} used contextual loss for style transfer and vector graphics generation.
Li~\etal\cite{li2020differentiable} demonstrated the usage of a differentiable renderer with a perceptual loss, to convert images into abstract SVGs.
Later, Liu~\etal\cite{liu2021paint} and Zou~\etal\cite{zou2021stylized} used a differential strokes renderer to produce a series of strokes for a given image, creating abstract paintings of the input image.
Vinker~\etal\cite{vinker2022clipasso, vinker2022clipascene} proposed a semantic CLIP loss for image-to-sketch translation, enhancing semantic fidelity while retaining morphology.
ClipVG~\cite{song2023clipvg} used differentiable rendering and CLIP-based optimization to edit images using vector graphics instead of pixels.
We follow~\cite{vinker2022clipascene}, demonstrating different levels of abstraction for vectorization tasks, by varying the number of shapes in the output image and by incorporating the suggested CLIP loss as a reconstruction objective.
\section{Method: Optimize \& Reduce}
\begin{figure}
\centering
\includegraphics[width=0.48\textwidth]{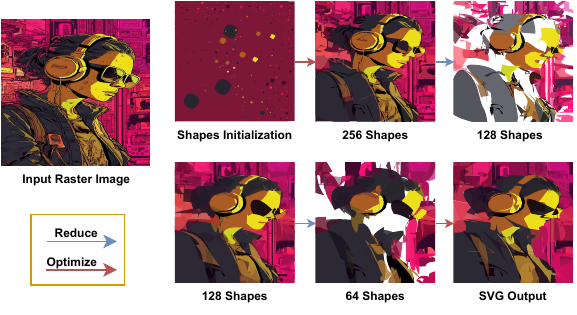}
\caption{\textbf{Algorithm Steps:} We initialize \besizer curves sampled uniformly on a circle with a radius inversely proportional to the DBSCAN clusters. We then iterate between optimization and reduction steps for the following schedule: $256\rightarrow128\rightarrow64$. The output SVG reconstructs the raster image with a small number of vector shapes.}
\label{fig:steps}
\end{figure}
\emph{Optimize \& Reduce} (O\&R) is a top-down approach to vectorization that is based on the DiffVG~\cite{li2020differentiable} differentiable renderer. Figures~\ref{fig:scheme} and~\ref{fig:steps} shows the core steps of our method. We start with a raster image, initialize the shapes and then iterate between Optimize and Reduce. For this scheme to work, we need to define how to initialize the shape parameters, how to determine which shapes to remove, and what loss function to use for DiffVG.

\paragraph{\bf Initialization}
We found that DiffVG is sensitive to the \Besizer curves initialization. Therefore, instead of initializing it with a random set of shape parameters, we use DBSCAN~\cite{ester1996density} to cluster image colors and divide the image into regions that correspond to connected components of each cluster. Then, we sort the connected components according to their pixel area in the source image. The $N$ largest connected components' center of mass serves as anchors for the differentiable rasterizer. The first optimization step of our method starts with circle-like shapes with a diameter proportional to the connected component area.

\paragraph{\bf Optimize}

The Optimize step runs DiffVG with the initial guess provided either by the initialization step or the Reduce step (described next). DiffVG takes as input a raster image as well the initial guess of $N$ shapes and optimizes the parameters of all the shapes {\em at once} to minimize a loss function between the resulting vector image and input raster image.

DiffVG is flexible with the choice of the optimization loss function $\mathcal{L}_{\text{optimize}}$, and we evaluate a number of loss functions. These can be any classical reconstruction pixel-wise loss, such as $L_1$ or $L_2$ (MSE) loss, as well as a contextualized loss, such as CLIP loss~\cite{vinker2022clipasso}. This loss uses different intermediate layers of CLIP-ViT~\cite{dosovitskiy2021image} where shallow layers preserve the geometry of the image and deeper layers preserve semantics.
We also devise a new geometric loss that regularizes the behavior of the \Besizer curves, explained later in this paper.
Combining all losses results in our general loss term:
\begin{equation}
\mathcal{L}_{optimize} = \mathcal{L}_{reconstruction} +  \lambda_{geometric}\mathcal{L}_{geometric}
\end{equation}

\paragraph{\bf Reduce}
A shape $P$ is defined via a set of \Besizer control points $p^{k}$ and an RGBA vector $c^{i}$. An $N$-shaped vector image is a set of $N$ shapes: $\Big \{ P_{i} = \{p_{i}^{k}, c_{i}^{j}\} \Big \} _{i=1}^{N}$. For each shape $P_{i}$, the ranking loss measures the degradation caused by not including that specific shape when the image is rendered. The loss is measured against the original raster image $\mathcal{I}$. Therefore, the loss for the $i$-th shape, includes rendering all shapes but the $i$-th shape:
\begin{equation}
    \hat{I} = \text{render}(\mathrm{P \setminus P_{i}})
\end{equation}
\begin{equation}
            \text{rank-score[i]} = \mathcal{L}_{\text{reduce}}(I, \hat{I})
\end{equation}

$\mathcal{L}_{\text{reduce}}$ can be any classical reconstruction pixel-wise loss, such as $L_{1}$ or MSE loss, a contextualized loss, or a non-differentiable loss, such as histogram matching loss. After scoring all shapes, the lower-ranked half of the shapes are removed from the vector representation of the image. The parameters of the remaining shapes serve as the initial guess for the {\em Optimization} step.

\paragraph{\bf Geometric Loss Term}
\begin{figure}
  \center
  \begin{tabular}{cccc}
    \begin{tabular}{l}
    \includegraphics[width=0.12\textwidth]{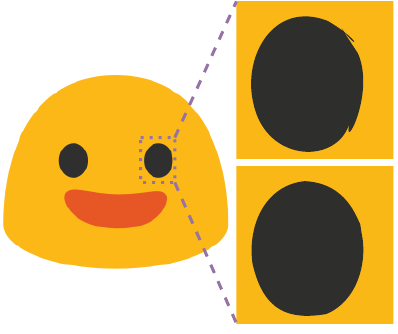}
    \end{tabular}
    & 
    \begin{tabular}{c}
    \includegraphics[width=0.06\textwidth]{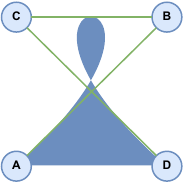}
    \end{tabular}
    & 
    \begin{tabular}{c}
    \includegraphics[width=0.06\textwidth]{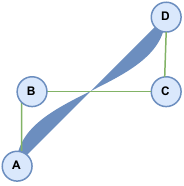}
    \end{tabular}
    & 
    \begin{tabular}{c}
    \includegraphics[width=0.06\textwidth]{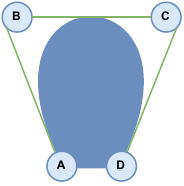}
    \end{tabular}
    \\
    (a) & (b) & (c) & (d)
    \end{tabular}  
  \caption{\textbf{Geometric Loss:} Our geometric constraint prevents curve intersections, resulting in the smoother topology of shapes. (a) Input image and SVG output without/with (Top/Bottom) geometric regularization. 
  Our geometric loss term has three components, each one aiming to eliminate intersection cases: (b) Self-intersection occurs when AB intersects CD. In addition, Cross-intersections are prone to happen when: (c)  $[A, B, C]$ and $[B, C, D]$ have different orientations. (d) $\angle ABC$ or $\angle BCD$ are acute.}
  \label{fig:geometric_all}
\end{figure}
Another contribution of our paper is a geometric loss term that encourages simple shapes through a differentiable penalty for intersections. An example in Figure~\ref{fig:geometric_all} shows the effectiveness of the proposed geometric loss in mitigating intersection problems.
The constraints on the control points prevent the circle shape from intersecting, demonstrating the intuitive and simple, yet highly effective nature of the proposed objective function.

The mathematical formulation is as follows: assume the control points of a cubic \Besizer path are (in order): A, B, C, and D. 
There are two types of intersections: self-intersections, where a single \Besizer curve intersects itself, and cross-intersections, where two \Besizer curves intersect each other within the same shape.
We observed intersections in three scenarios: (a) Self-intersection exists if the lines $\vec{AB}$ and $\vec{CD}$ intersect (Figure~\ref{fig:geometric_all}b). Furthermore, cross-intersections \textit{are prone to happen} when: (b) ABC and BCD have different orientations (Figure~\ref{fig:geometric_all}c), (c) $\angle ABC$ or $\angle BCD$ are acute (Figure~\ref{fig:geometric_all}d).

To measure self-intersection or orientation change, we compute the orientation (0 is clockwise and 1 is counter-clockwise) of 3 points $[A, B, C]$, where $S(x)$ represents a sigmoid function: 
\begin{equation}
\begin{split}
    O(A,B,C) = S(&(B_y - A_y) \cdot (C_x - B_x) - \\
                 &(B_x - A_x) \cdot (C_y - B_y))
\end{split}
    \label{eq:orientation}
\end{equation}
Using this approximation and simple AND and XOR logic, we can verify the same orientation for $[A, B, C]$ and $[B, C, D]$, or check whether AB intersects CD ($f_{intersect}$ and $f_{orientation}$ defined and detailed in the supplementary).
We integrate these functions into our geometric loss:
\begin{equation}
\mathcal{L}_{geom_{ab}} = \lambda_{p} \cdot (f_{intersect} + f_{orientation})
\end{equation}
In addition, The angle loss can be computed using a dot product between the line segments:
\begin{equation}
\begin{split}
\mathcal{L}_{geom_{c}} = &ReLU(-\frac{AB \cdot BC}{|AB| \cdot |BC|}) + \\
                         &ReLU(-\frac{BC \cdot CD}{|BC| \cdot |CD|})
\end{split}
\end{equation}

Our final geometric loss is:
\begin{equation}
\mathcal{L}_{geometric} = \mathcal{L}_{geom_{ab}} + \mathcal{L}_{geom_{c}}
\end{equation}

LIVE~\cite{ma2022towards} suggests a different geometric loss, that encourages the angle between the first $\vec{AB}$ and the last $\vec{CD}$ control points connections to be greater than $180^{\circ}$. Though this rule encourages shapes to be convex, our loss term also avoids both self-intersection and orientation changes.

\section{Experiments}
\begin{figure}
  \centering
  \begin{tabular}{ccccc}
    \includegraphics[width=0.07\textwidth]{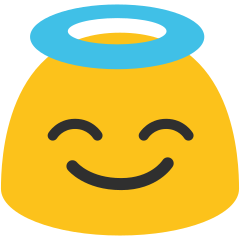} & 
    \includegraphics[width=0.07\textwidth]{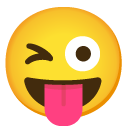} & 
    \includegraphics[width=0.083\textwidth]{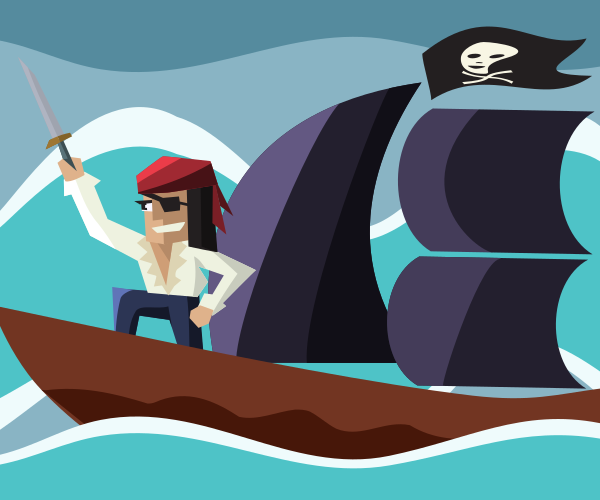} &
    \includegraphics[width=0.07\textwidth]{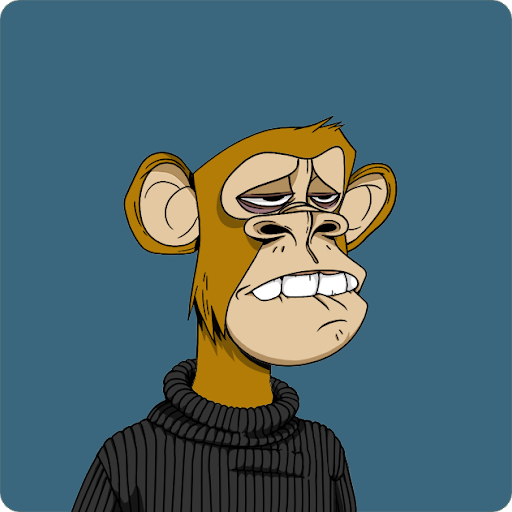} & 
    \includegraphics[width=0.07\textwidth]{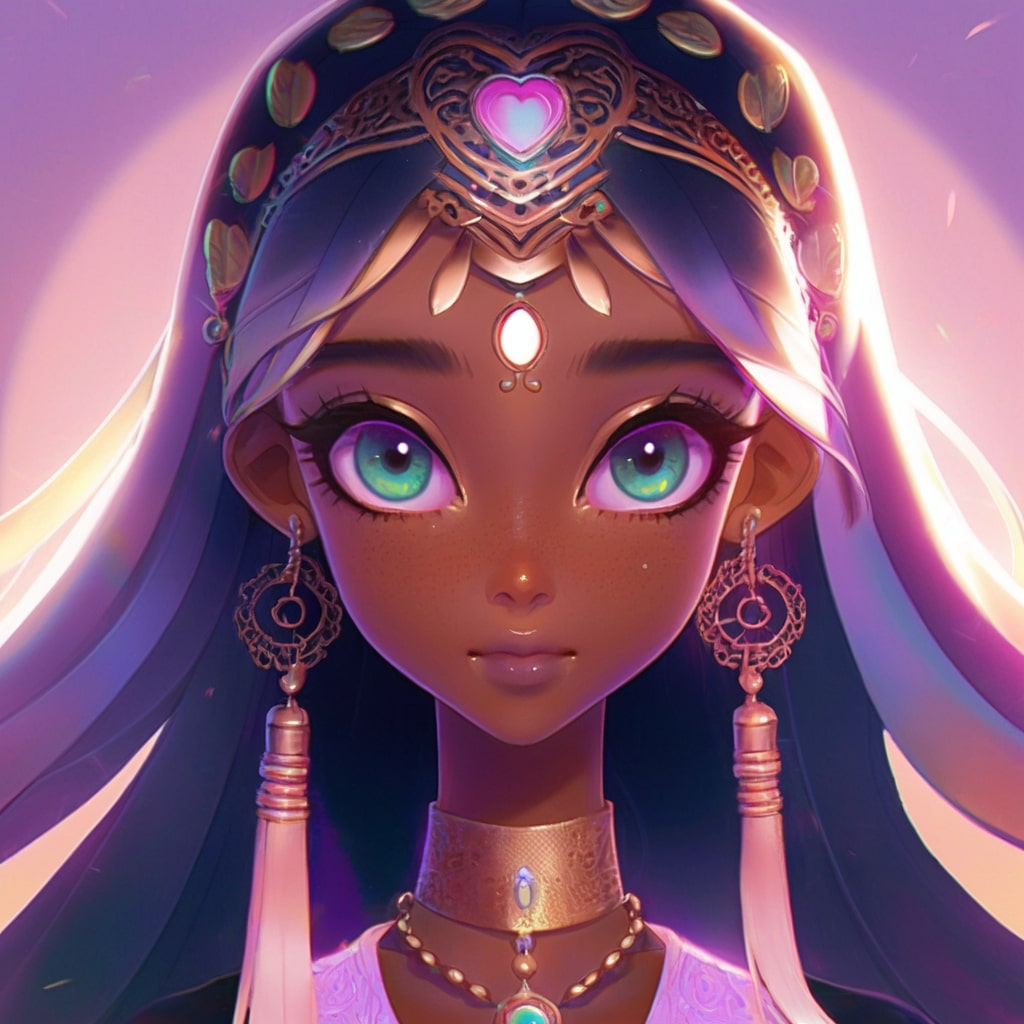} \\
     (a) & (b) & (c) & (d) & (e)
  \end{tabular}
  \caption{\textbf{Dataset Visualization:} We evaluate our algorithm on five datasets: (a) former version of EMOJI~\cite{notoemoji} (b) current version of EMOJI~\cite{notoemoji} (c) images gathered from free-svg website (d) images from the famous bored apes NFTs (e) images generated using a text-to-image model (Midjourney).}
  \label{fig:dataset_vis}
\end{figure}

We introduce five new datasets and compare our method against prior art. O\&R is a single-image optimization method, therefore we compare it with single-image methods: DiffVG~\cite{li2020differentiable} and LIVE~\cite{ma2022towards}. We then demonstrate applications based on our method.

\subsection{Datasets}
For evaluation, we collect five datasets to cover a range of image complexities, including emojis, cliparts, and natural-like images. Figure~\ref{fig:dataset_vis} displays a sample image from each dataset. All datasets will be made publicly available.

\textbf{EMOJI dataset}. We take two snapshots of the NotoEmoji project \cite{notoemoji}. 
The first, taken in 2015, follows LIVE's EMOJI's dataset. The second from 2022 introduces color gradients and is available in various resolutions. For evaluation, we resize all of the acquired images to a $240 \times 240$ resolution. The supplemental includes a list of all emojis' names and a showcase of the two datasets.

\textbf{Free-SVG}. A collection of 65 images, including complex clipart images. Compared to the Emoji dataset, this dataset is more complex and difficult for image vectorization.

\textbf{NFT-Apes}. A collection of 100 images from the famous Bored Apes Yacht Club NFT. It features profile pictures of cartoon apes that are generated by an algorithm. It includes images with various backgrounds and complex shapes.

\textbf{Text-To-Image Generated Images}. To stress-test vectorization algorithms we introduce a dataset of images generated by Midjourney, a Text-To-Image public generator. We gather 100 such images alongside their prompts.

\subsection{Implementation Details}
\begin{figure*}
  \centering
  \begin{tabular}{ccccc}
  Old Emojis & New Emojis & Free-SVG & NFT-Apes dataset & Midjourney \\
        \includegraphics[width=0.18\textwidth]{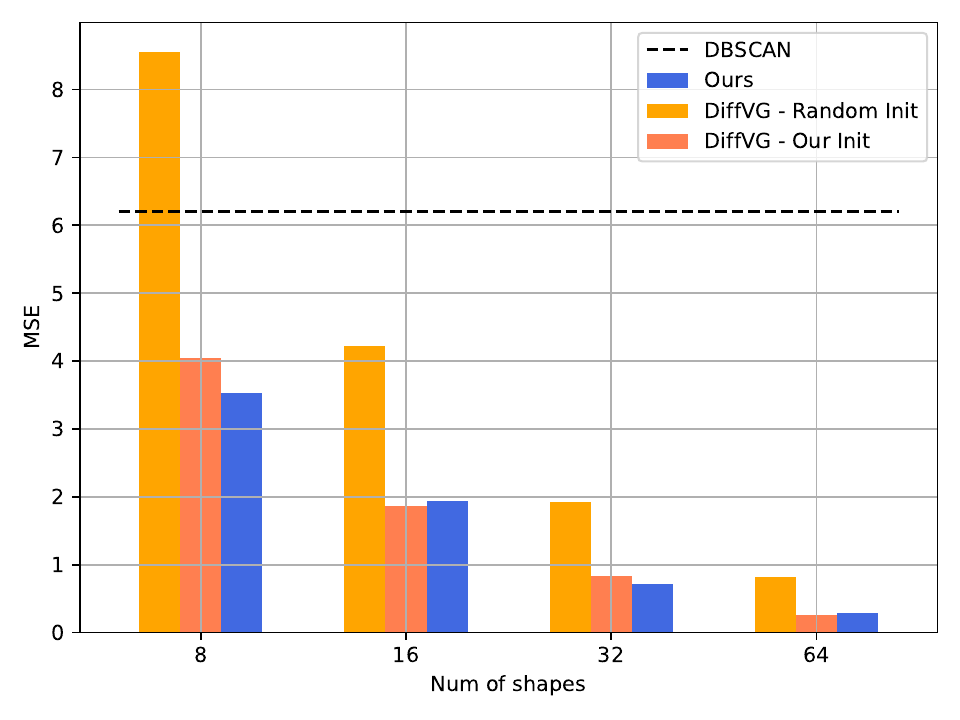} &
    \includegraphics[width=0.18\textwidth]{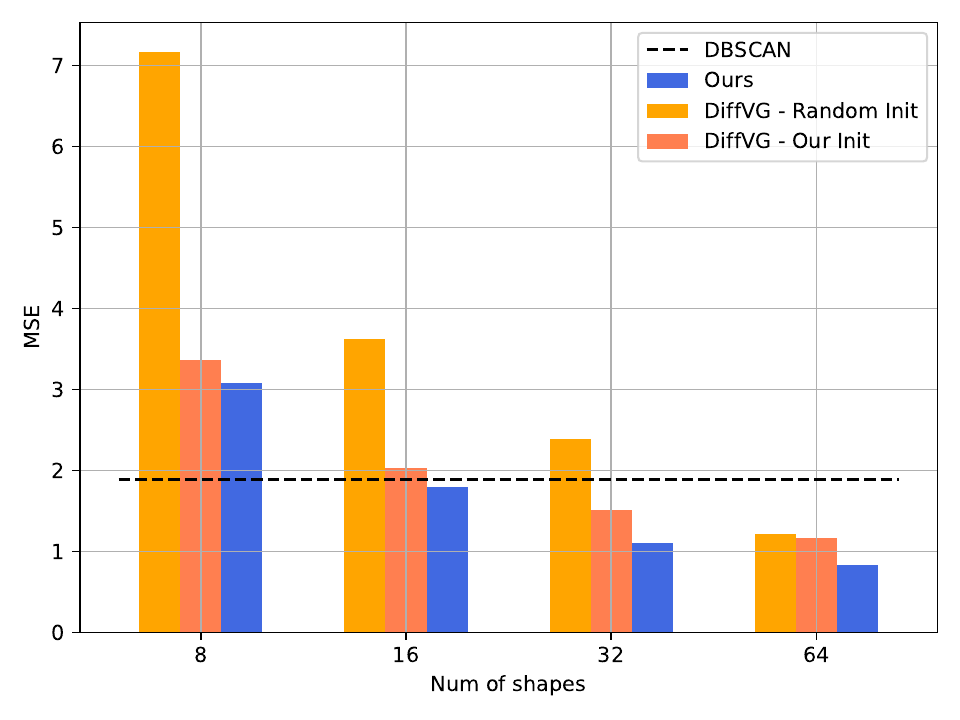} &
    \includegraphics[width=0.18\textwidth]{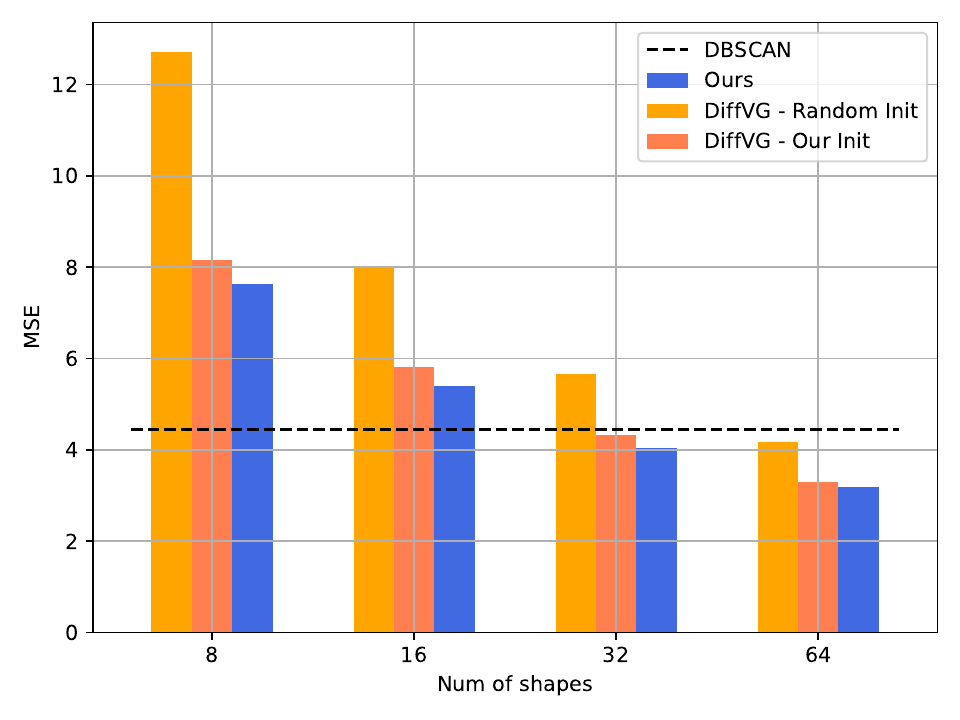} & 
    \includegraphics[width=0.18\textwidth]{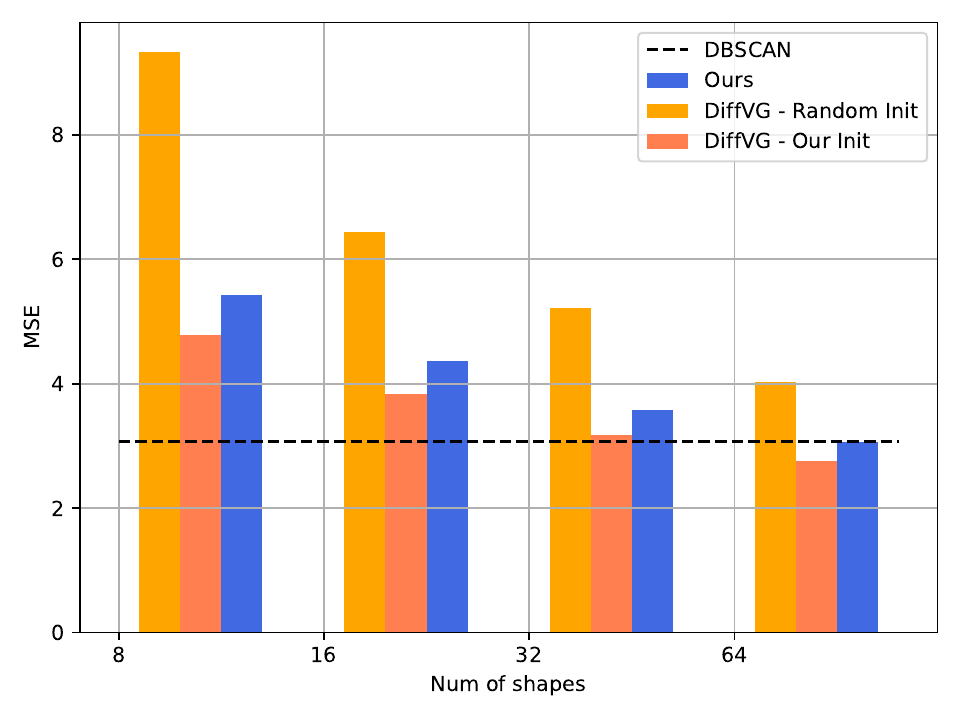} &
    \includegraphics[width=0.18\textwidth]{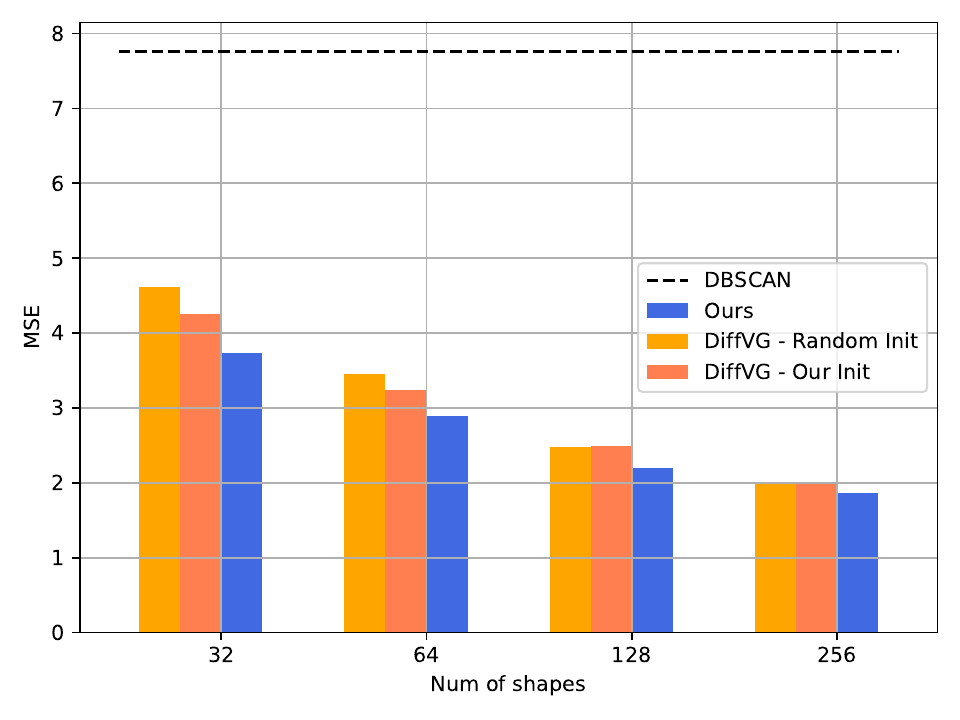} \\
    \end{tabular}
  \label{tab:my_table}

\caption{\textbf{MSE ($\downarrow$ better) vs Number of shapes:} Our method (blue) outperforms DiffVG with random  initialization (orange) for all five datasets. Moreover, using our proposed initialization (peach) improves the performance of DiffVG. Dashed line represents the MSE between DBSCAN-clustering and input image. 
}
\label{fig:mse_vs_num_shapes}
\end{figure*}

Our code is written in PyTorch~\cite{paszke2017automatic}. Hyperparameters and baseline implementations are described thoroughly in the supplementary.

We use three Reduce steps that follow an exponential decay schedule (i.e., the number of shapes is halved in each iteration). 
For the qualitative experiments, we use MSE for optimization and $L_{1}$ for reduction.
All quantitative evaluations use $L_{1}$ with our geometric loss for optimization.
For reduction, we use $L_{1}$ and Clip on Midjourney, and for the other datasets, we use only $L_{1}$.
Further details on the runtime analysis are in the supplementary.

\subsection{Quantitative Evaluation}
Figure~\ref{fig:live_with_runtime_after_refactor} shows MSE reconstruction error vs the number of vector shapes for the new emoji dataset. Colors in the graph code DiffVG, LIVE, and our method. As can be seen, DiffVG is the fastest but converges to solutions that are not very accurate (especially when the number of shapes is low). LIVE, on the other hand, works well, but takes about 30 minutes\footnote{Runtime was measured on an NVIDIA RTX A5000 GPU.} on average to vectorize a 64 shape image. O\&R strikes a good balance between the two. It is almost two orders of magnitude faster than LIVE and as accurate. The speedup is a result of our exponential scheduler. For $n$ shapes, LIVE optimizes layer-by-layer which results in $\mathcal{O}(n)$ steps, while we optimize $\mathcal{O}(1)$ steps. For precise runtime figures, please refer to the supplementary materials.

Figure~\ref{fig:mse_vs_num_shapes} displays the MSE reconstruction error plotted against the number of shapes in the vector representation. 
Across all datasets, our method consistently outperforms DiffVG with random initialization, when considering the same number of shapes. This observation holds true for the LPIPS score (using VGG~\cite{simonyan2014very}) plotted against the number of shapes, which is presented in the supplemental.
It also outperforms DiffVG with our suggested initialization, except for the NFT-Apes dataset, where our DBSCAN initialization introduces a strong inductive bias, resulting in improved performance of DiffVG when using our initialization.

\subsection{Qualitative Evaluation}

\begin{figure}
  \begin{tabular}{c|cccc}
    \includegraphics[width=0.065\textwidth]{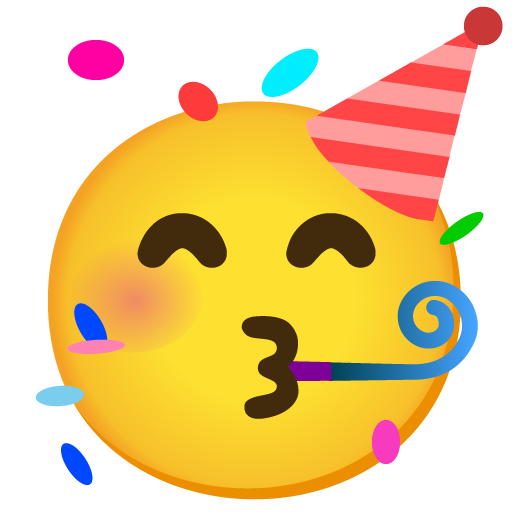} & 
    \includegraphics[width=0.065\textwidth]{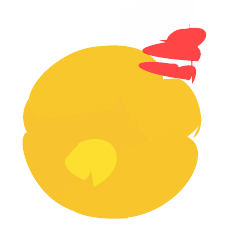} & 
    \includegraphics[width=0.065\textwidth]{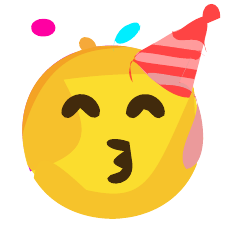} &
    \includegraphics[width=0.065\textwidth]{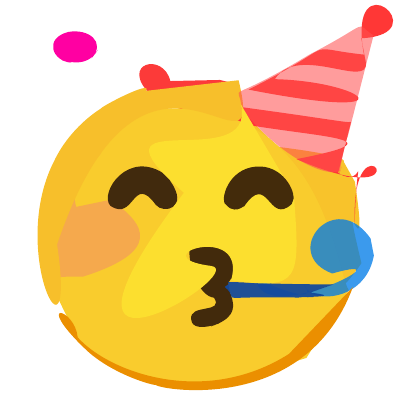} &
    \includegraphics[width=0.065\textwidth]{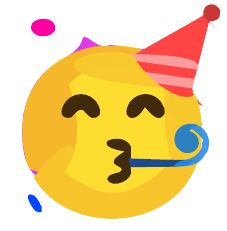}\\
     Raster & DiffVG & DiffVG & \textbf{Ours} & \textbf{Ours} \\
     Image & Rand & DBSCAN & Rand & DBSCAN \\
     (a) & (b) & (c) & (d) & (e) \\

    \end{tabular}
  \caption{\textbf{Components Contribution:} Results using 16 shapes. DiffVG is remarkably sensitive to the curve's initialization, completely missing the party horn.}
  \label{fig:DBSCAN_contribution_emojis}
\end{figure}

\begin{figure}
  \centering
  \begin{tabular}{c|c}
    \includegraphics[width=0.2\textwidth]{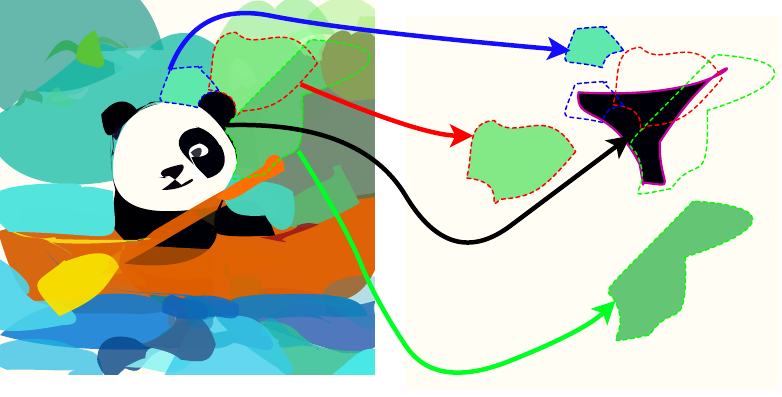} &
    \includegraphics[width=0.2\textwidth]{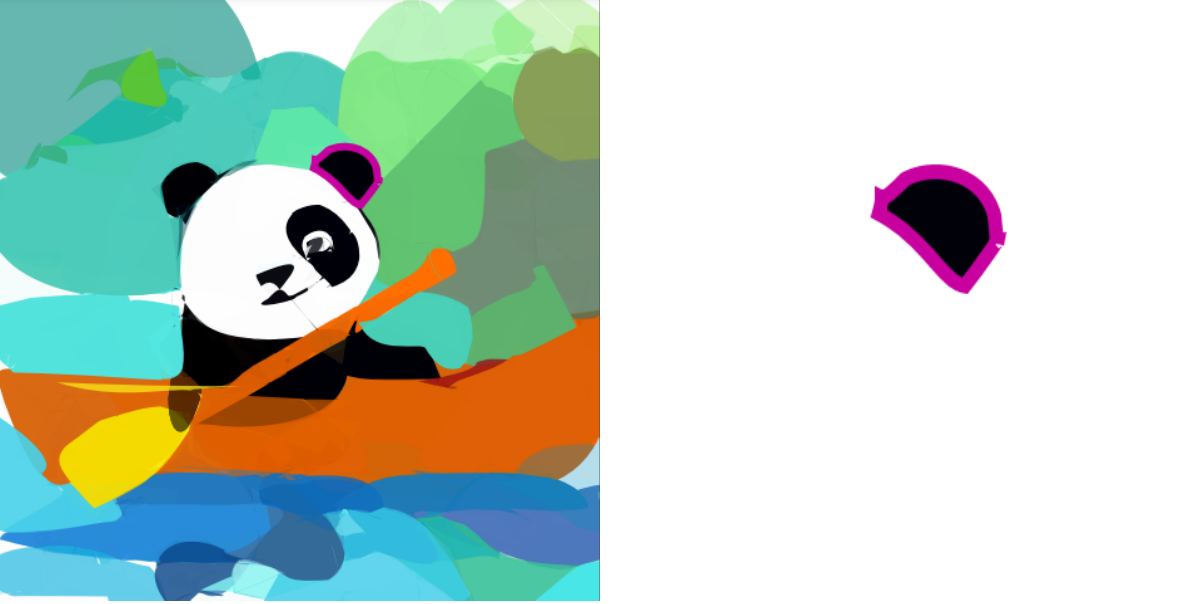} \\
    (a) VectorFusion  & (b) Ours \\
    
    \end{tabular}
  \caption{\textbf{Convex Shapes: VectorFusion vs Ours :} For VectorFusion, the right ear is composed of not less than four shapes. In our case, the ear is made of one shape. This demonstrates our compact representation and editability.}
  \label{fig:geometric_vectorfussion}
\end{figure}

\paragraph{The Contribution of DBSCAN and O\&R} 
Figure~\ref{fig:mse_vs_num_shapes} shows that DBSCAN initialization reduces DiffVG's error. However, when a low number of shapes is considered, the introduction of O\&R further decreases the error. Figure~\ref{fig:DBSCAN_contribution_emojis} provides insight into the effectiveness of our components - DBSCAN initialization and O\&R optimization scheme. Figures~\ref{fig:DBSCAN_contribution_emojis}(b) and (c) demonstrate DiffVG's sensitivity to initialization, emphasizing its limitations without effective initialization. Furthermore, the contribution of O\&R optimization in mitigating local minima is showcased (Figures~\ref{fig:DBSCAN_contribution_emojis}(c) and (d)). The comparison underscores that DBSCAN initialization alone falls short in yielding semantically satisfactory results for DiffVG. Notably, the primary driver of our algorithm's success lies in the O\&R top-down approach, as demonstrated in Figures~\ref{fig:DBSCAN_contribution_emojis}(d) and (e).

\begin{figure}
\centering
\includegraphics[width=0.48\textwidth]{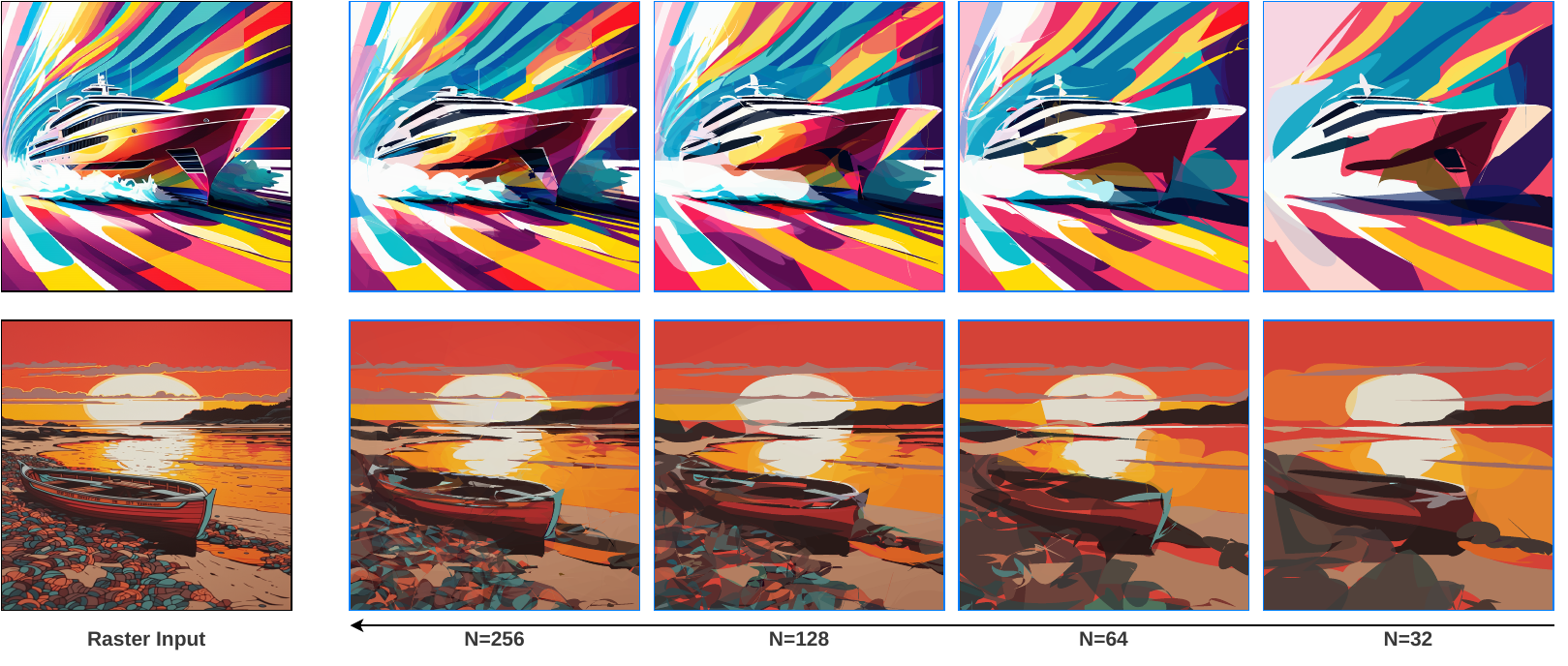}
\caption{\textbf{Results Showcase:} We showcase more results of our method on a collection of images generated using Midjourney. Each image is vectorized using 256, 128, 64, and 32 shapes.}
\label{fig:midjourney_more_results}
\end{figure}
\begin{figure}
\centering
\includegraphics[width=0.48\textwidth]{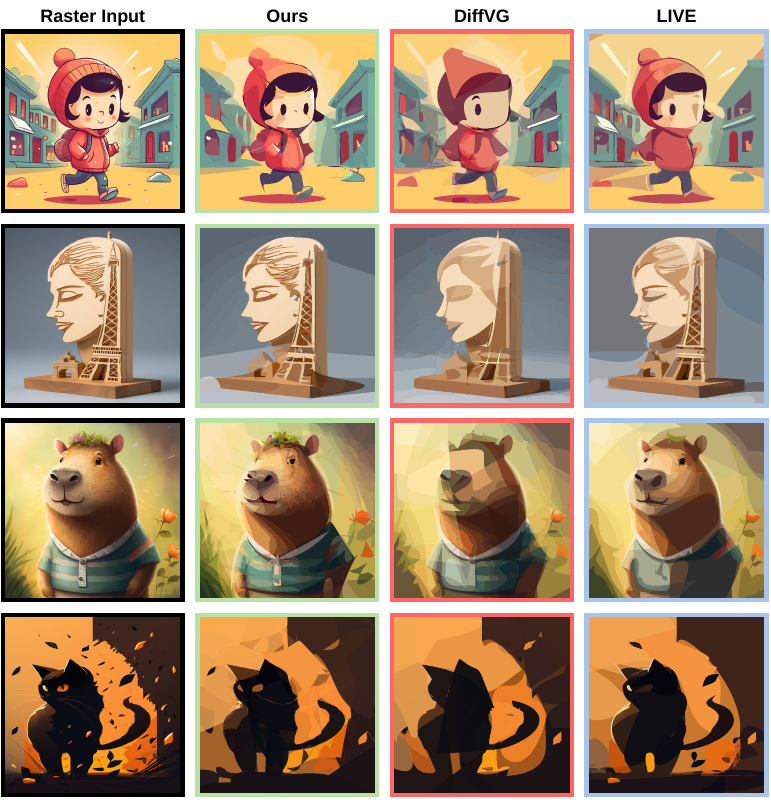}
\caption{\textbf{Qualitative Comparison:} We showcase the results of DiffVG, LIVE, and Our method on a collection of images generated using Midjourney. Each image is compared using the same number of shapes.}
\label{fig:comparison_midjourney}
\end{figure}
\paragraph{Geometric Loss}  VectorFusion~\cite{jain2022vectorfusion} is a Text-to-SVG algorithm that given an optimizer such as DiffVG~\cite{li2020differentiable} or LIVE~\cite{ma2022towards}, utilizes a text conditional diffusion model to create an SVG image from a text prompt. Figure~\ref{fig:geometric_vectorfussion} shows VectorFusion's output using LIVE's framework with its geometric loss and with SDS~\cite{poole2022dreamfusion} loss, for "A Panda rowing a boat" prompt. Figure~\ref{fig:geometric_vectorfussion}a shows LIVE's decomposition of the shapes composing the panda's ear. It takes four shapes to capture the ear's shape. Furthermore, the black shape is non-convex and exhibits a self-intersection. Figure~\ref{fig:geometric_vectorfussion}b shows the result of taking the image rasterizing it, and vectorizing it with O\&R. As highlighted with a pink solid line, we capture the panda's ear using a single \emph{convex} shape. 
Additional results and comparisons can be seen in Figures~~\ref{fig:midjourney_more_results} and \ref{fig:comparison_midjourney}.

\subsection{Applications}
\paragraph{\bf Image Abstraction}
\begin{figure}
\centering
\includegraphics[width=0.4\textwidth]{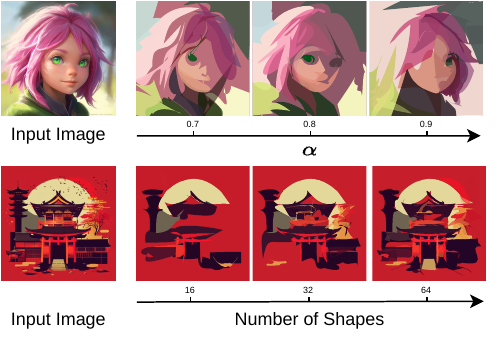}
\caption{\textbf{Abstraction Examples:} Examples of (Top) different levels of abstraction using 16 shapes and various $\alpha$ values and number of shapes. Lower $\alpha$ means more abstraction (Equation~\ref{eq:abstraction}). (Bottom) various number of shapes.}
\label{fig:clip_matrix2}
\end{figure}
Inspired by CLIPasso~\cite{vinker2022clipasso}, we demonstrate image abstraction with two degrees of freedom: (1) the number of shapes in the vector image representation, and (2) the extent to which we utilize a CLIP loss for optimization. Accompanied by the Midjourney image prompts, we propose employing CLIP loss, incorporating both its image-to-image components (image embeddings and image attention maps) and text-to-image components.

We define a hyperparameter $\alpha$ in Equation~\ref{eq:abstraction}, which controls the extent to which a reconstruction loss ($L_{1}$ loss) is preferred over a contextualized loss ($\mathcal{L}_{CLIP}$) when optimizing for shapes parameters. The top row of Figure~\ref{fig:clip_matrix2} demonstrates the effect of $\alpha$ on O\&R for different values of $\alpha$. 
For small values of $\alpha$ the image retains semantic features, whereas for larger values details are neglected, reflecting the effect of $L_{1}$ loss.
\begin{equation}
\mathcal{L} = \alpha{L_{1}} + (1-\alpha)\mathcal{L}_{CLIP} 
    \label{eq:abstraction}
\end{equation}
Alternatively, we achieve abstraction with a lower number of shapes as demonstrated in the bottom row of Figure~\ref{fig:clip_matrix2}. Despite the significant variation in abstraction level, the output SVG remains easily identifiable as depictions of the same input image.

\paragraph{\bf Interpolation} 

Previous works, such as Im2Vec~\cite{reddy2021im2vec} and LIVE~\cite{ma2022towards}, have proposed interpolating emojis and character images in latent space, producing an interpolated latent representation that serves as input to a GAN, which generates a raster image. This raster image is then vectorized to create the vector output. In contrast, our approach does not rely on a GAN to generate interpolated images; instead, it performs interpolation directly in the shapes domain between the source and target images. Our method is preferable as it reflects a realistic scenario where a user wants to interpolate between two emoji images they possess. Our results are achieved without depending on a GAN trained to generate the raster images.

\begin{figure}
  \centering
  \begin{tabular}{c}
     \includegraphics[width=0.48\textwidth]{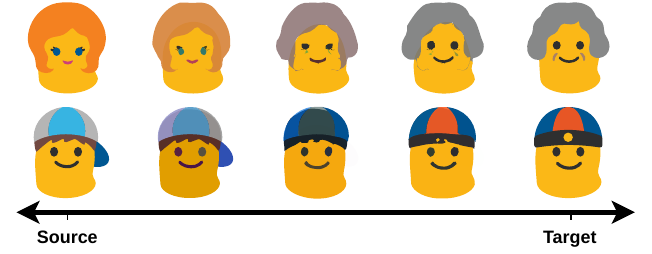} 
  \end{tabular}
  \caption{\textbf{Interpolation:} We perform an interpolation of emoji images from the rightmost column to the leftmost column. 
  The intermediate images shown in between represent the optimization process's incremental results.}
  \label{fig:interpolation_with_initialization}
  
\end{figure}
Figure~\ref{fig:interpolation_with_initialization} demonstrates an image interpolation application between two emoji images. 
Interpolating vector images imposes a double correspondence problem: (1) Resolving shape correspondence and (2) \Besizer control point correspondence. 
To overcome this problem, we propose a new mechanism. We start with vectorizing the first image using O\&R. Then, we use the vectorized image as the initial approximation for the shape structure and color and optimize the shapes, setting the second image as the target. During optimization, we obtain interpolated states between the two images. We leverage CLIP loss as an optimization objective for O\&R to semantically interpolate the two images.

\section{Conclusions}
Optimize \& Reduce is a top-down approach for vectorization, which excels when the budget of the number of shapes is low. 
The combination of O\&R steps help the algorithm avoid local minima and converge to a solution that minimizes the reconstruction loss. Since O\&R is a top-down approach, its' runtime is much faster than bottom-up approaches. Furthermore, our DBSCAN-based initialization and geometric loss term results in fast and editable results. 

As part of our contribution, we suggest a benchmark dataset that facilitates the standardization of vectorization methods. We supplement our demonstrated reconstruction capabilities on this benchmark with applications of image abstraction and emoji interpolation. 

\paragraph{Acknowledgments:} Parts of this research were supported by ISF grant 1549/19.

% \section{Acknowledgments}
% Thanks
\bibliography{aaai24}
\clearpage
\section{Supplementary}
\subsection{Optimize \& Reduce}
\begin{figure}
  \centering
  \begin{tabular}{cccc}
    \includegraphics[width=0.1\textwidth]{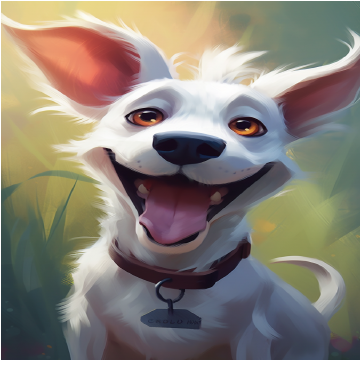} & 
    \includegraphics[width=0.1\textwidth]{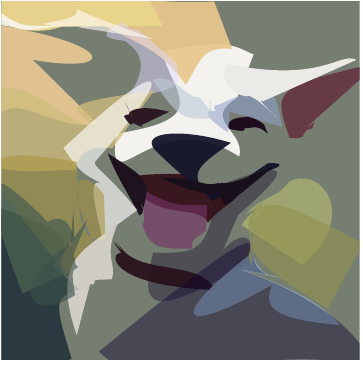} & 
    \includegraphics[width=0.1\textwidth]{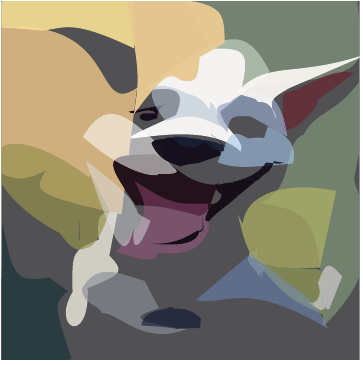} & 
    \includegraphics[width=0.1\textwidth]{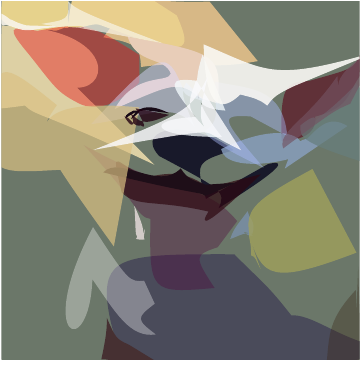} \\
    Original & \multicolumn{3}{c}{Stochastic samples}
    
    \end{tabular}
  \caption{\textbf{Reduction Sampling:} Results of different runs for the same input configuration, due to the stochastic nature of our reduce step. We reduce the number of shapes drastically to emphasize the effect of the sampling.}
  \label{fig:sample_same}
\end{figure}

\begin{figure}
  \centering
    \begin{tabular}{ccccc}
    $N=64$ & \textcolor{red}{$64\rightarrow32$} & $N=32$ & \textcolor{cyan}{$32\rightarrow64$} & $N=64$ \\
    \includegraphics[width=0.07\textwidth]{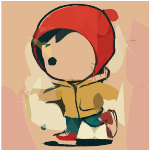} & 
    \includegraphics[width=0.07\textwidth]{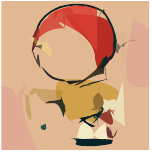} & 
    \includegraphics[width=0.07\textwidth]{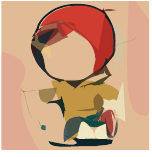} & 
    \includegraphics[width=0.07\textwidth]{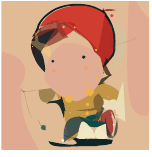} &
    \includegraphics[width=0.07\textwidth]{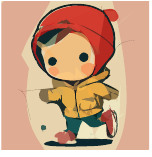} \\
    (a) & (b) & (c) & (d) & (e) \\
    
    \end{tabular}
  \caption{\textbf{Add Operation:} We run O\&R with a schedule of:
  $256\!\rightarrow\!128\!\rightarrow\!64\!\rightarrow\!32\!\rightarrow\!64$ shapes. (a) illustrates the Optimization process to $64$ shapes, (b) demonstrates the Reduce step to $32$ shapes, (c) displays the optimization of $32$ shapes, (d) shows the initialization of $32$ additional \Besizer shapes, and (e) exhibits the result of the Add operation, bringing the shape count back to $64$. Once the foundation of $32$ shapes is established, adding $32$ shapes captures previously missing details like the subject's eyes and cheek blush.}
  \label{fig:add_example}
\end{figure}

\begin{figure}
  \centering
  \begin{tabular}{ccc}
    \includegraphics[width=0.1\textwidth]{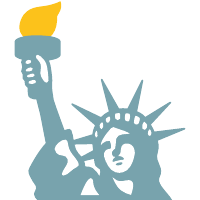} &
    \includegraphics[width=0.1\textwidth]{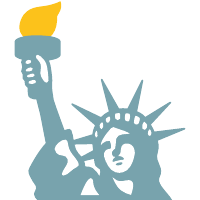}  &
    \includegraphics[width=0.1\textwidth]{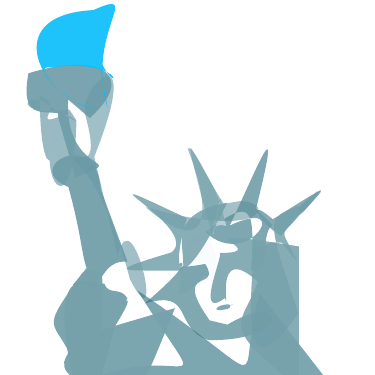} \\
    (a) & (b) & (c) \\
       \includegraphics[width=0.1\textwidth]{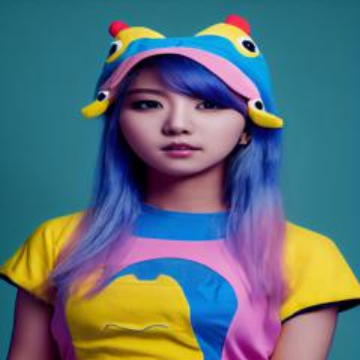}  &
    \includegraphics[width=0.1\textwidth]{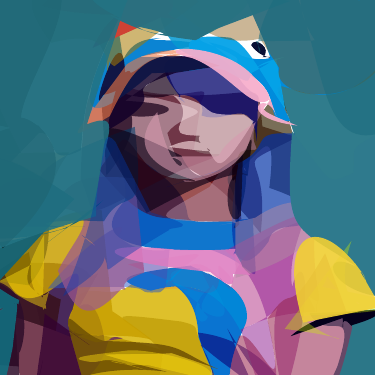}  & 
    \includegraphics[width=0.1\textwidth]{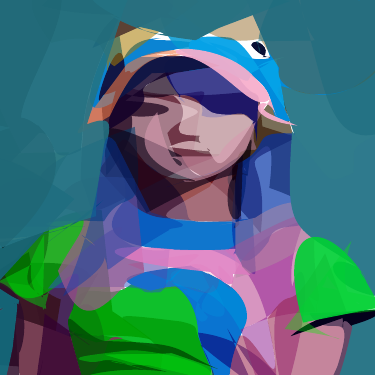} \\
    (d) & (e) & (f) \\
  \end{tabular}
  \caption{\textbf{Color Editing}. We use O\&R to vectorize the raster images: (a) / (d) to (b) / (e) respectively. Subsequently, we perform color edits on the \Besizer shapes to obtain the color-edited images shown in (c) / (f).}
  \label{fig:qualitative_editing_color}
\end{figure}

\paragraph{\textbf{Optimize \& Reduce stated formally:}} Algorithm~\ref{alg:algorithm} provides a comprehensive outline of the Optimize \& Reduce (O\&R) algorithm. We define explicit functions for the two steps of our method and describe how they are orchestrated to achieve our goal. The algorithm begins with a raster image and utilizes DBSCAN for initialization. We then iteratively perform the Optimize and Reduce steps. 
\RestyleAlgo{ruled}
\begin{algorithm}
    \caption{Algorithm of Optimize \& Reduce}
    \label{alg:algorithm}
    \KwIn{I, shapes, iters}
    \KwOut{SVG}
    
    \SetKwFunction{optimize}{Optimize}
    \SetKwProg{Fn}{Function}{:}{}
    \Fn{\optimize{I, P}}{
        \For{$j=1$ to t}{\
            $\hat{I} = \mathrm{render}(\mathrm{P})$\;
            $\mathcal{L} = \mathcal{L}_{\text{optimize}}(I, \hat{I}, P)$ \;
            $\mathrm{P} = P-\alpha \frac{d\mathcal{L}(\mathrm{P})}{d\mathrm{P}}$ \textcolor{gray}{\tcp*{update shapes}}
        }
    \textbf{return} $ P ;$
    }
    \SetKwFunction{reduce}{Reduce}
    \SetKwProg{Fn}{Function}{:}{}
    \Fn{\reduce{I, P, n}}{
        $L = [] $ \;
        \For{$j=1$ to len(P)}{
            $\hat{I} = \mathrm{render}(\mathrm{P \setminus P[j]})$ \;
            $loss = \mathcal{L}_{\text{reduce}}(I, \hat{I}, P)$ \;
            $L = concat([L;\quad loss])$ \;
        }
        prob = SoftMax(L) \;
        P = Sample(P, prob, n) \textcolor{gray}{\tcp*{sample n shapes}} 
    \textbf{return} $ P ;$
    }
    $P = \mathrm{init}(I, 2^{iters} \cdot shapes)$ \;
    $P = \optimize{I, P} $\;
    \For{$n$ in [$2^{iters-1} \cdot \text{shapes}, ..., 2 \cdot \text{shapes}, \text{shapes}$] }{ 
        P = \reduce{I, P, n} \;
        P = \optimize{I, P} \;
    }
\end{algorithm}
\paragraph{\bf Stochastic Reduce}
Instead of deterministically removing the bottom half of shapes, we can use sampling to introduce variability. We convert the rank scores to a probability distribution with SoftMax:
\begin{equation}
\text{SoftMax}(z)_i = \frac{e^{z_{i}}/T}{\sum_{j=1}^N e^{z_{j}/T}} \ \ \ for\ i=1,2,\dots,N
\end{equation}
We control the extent of variability through the hyperparameter $T$. Smaller values of $T$ bias the SoftMax operation towards deterministic sampling, where the top-k most important shapes are selected. Conversely, larger values of $T$ lead to sampling under a uniform distribution regime. Figure~\ref{fig:sample_same} shows examples of different results created using the same stochastic reduce.
\RestyleAlgo{ruled}
\begin{algorithm}
    \caption{Probabilistic Reduce}
    \label{alg:probabilistic_reduce}
        \SetKwFunction{reduce}{Probabilistic-Reduce}
    \SetKwProg{Fn}{Function}{:}{}
    \Fn{\reduce{I, P, n}}{
        $\text{shape-score} = [] $ \;
        \For{$j=1$ to len(P)}{
            $\hat{I} = \mathrm{render}(\mathrm{P \setminus P[j]})$ \;
            $\text{shape-score}[j] \leftarrow \mathcal{L}_{\text{reduce}}(I, \hat{I})$ \;
        }
        prob = SoftMax(L) \;
        P = Sample(P, prob, n) \textcolor{gray}{\tcp*{sample n shapes}} \
    \textbf{return} $P$ \;
    }
\end{algorithm}
\begin{figure}
  \centering
  
  \begin{tabular}{ccc}
    \includegraphics[width=0.13\textwidth]{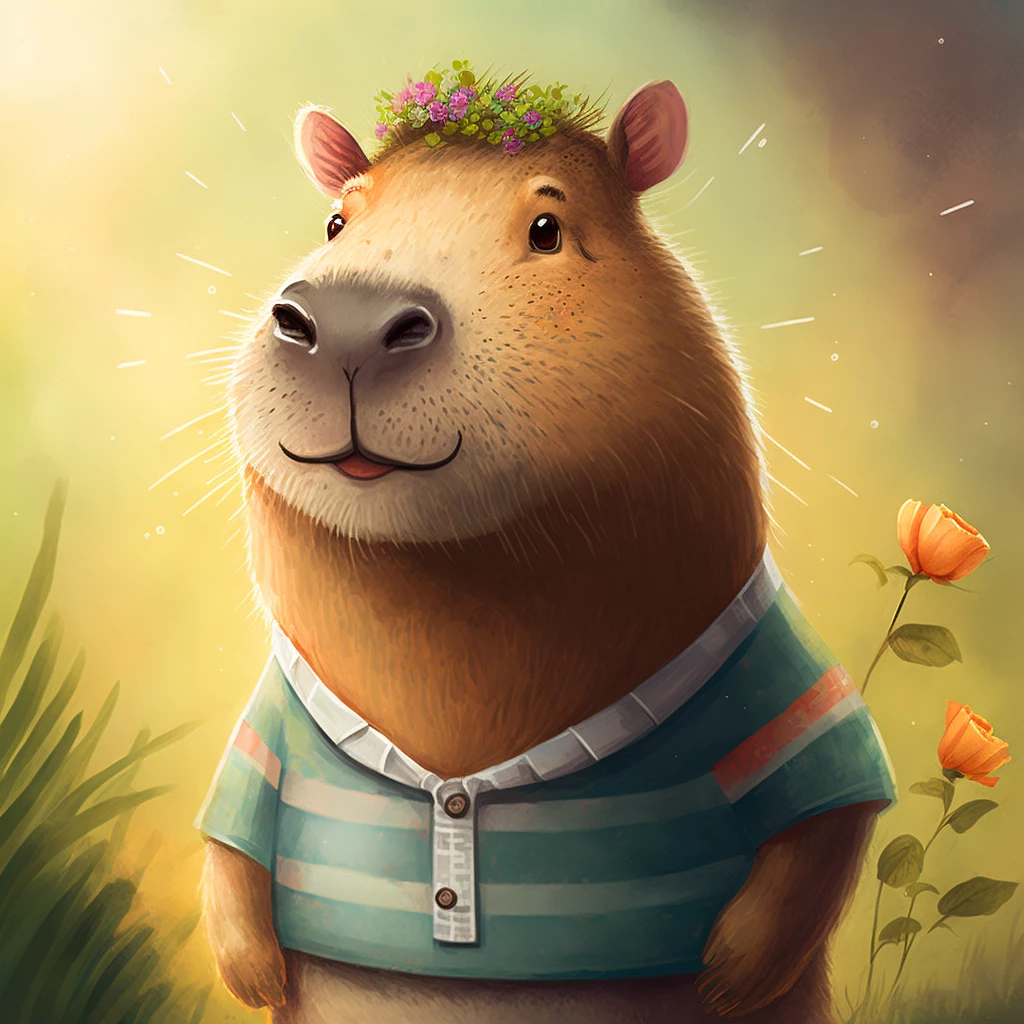} &
    \includegraphics[width=0.13\textwidth]{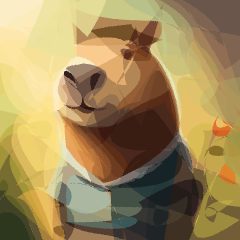} &
    \includegraphics[width=0.13\textwidth]{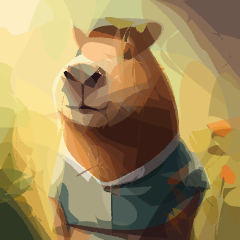} \\
    (a) Original & (b) $T=5\cdot10^{-4}$ & (c) $T=1\cdot10^{-3}$ \\
    \includegraphics[width=0.13\textwidth]{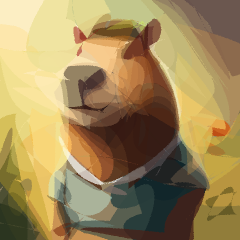} &
    \includegraphics[width=0.13\textwidth]{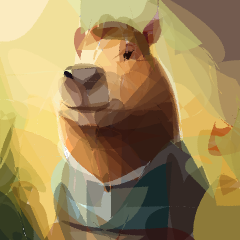} & 
    \includegraphics[width=0.13\textwidth]{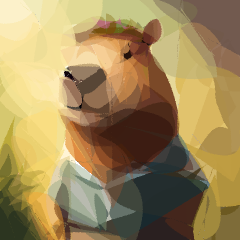} \\
     (d) $T=0.01$ & (e) $T=0.1$ & (f) $T=1$
    \end{tabular}
    
  \caption{\textbf{Sampling Temperature:} Results using different SoftMax temperatures. Lower temperature means more deterministic sampling}
  \label{fig:sampling_temp}
\end{figure}
\paragraph{\bf Adding Shapes} O\&R leverages DiffVG to optimize shape parameters and incorporates a Reduce step to avoid local minima. We employ an exponential Reduce scheduler that \emph{halves} the number of \Besizer shapes in each iteration. 
However, for images with intricate details or prominent textures, a low number of shapes leads to coarse segmentation. To enhance the representation of textures and preserve fine details, we introduce an {\em Add} operation. Inspired by LIVE~\cite{ma2022towards}, we selectively add shapes to regions with high reconstruction error and optimize these newly added shapes, all at once, using DiffVG. Figure~\ref{fig:add_example} illustrates how the add operation boosts performance. Adding $32$ shapes back after O\&R to $32$ shapes adds back the subjects' eyes and cheek blush. 
\paragraph{\textbf{Initialization:}} We emphasize the importance of a robust initialization method for initializing Bézier shapes, as it plays a vital role in the differentiable rendering-based image vectorization process. Our initialization technique is based on DBSCAN, a clustering algorithm that does not require a prior assumption on the number of clusters. In the main paper, we compare the Mean Squared Error (MSE) with the number of shapes for DiffVG using random initialization, DiffVG using our initialization, and Our method. Through this graph, we quantitatively demonstrate our dual contribution, showcasing consistent improvement from DiffVG with random initialization to DiffVG with our initialization. Additionally, for all datasets except NFT-Apes, transitioning from DiffVG to O\&R as the optimization method, leads to further enhancements in reconstruction results. 
For a qualitative assessment, we direct the reader to Figure~\ref{fig:DBSCAN_contribution_emojis}, which illustrates the impact of our initialization. Without our initialization, DiffVG with random initialization fails to capture important shapes in the image such as the mouth and eyes. Our initialization aids in recovering the mouth region, while with O\&R, with the same final shape budget, we successfully reconstruct all facial attributes of the emoji.
\begin{figure}
  \centering
  \begin{tabular}{cc}
    \includegraphics[width=0.1\textwidth]{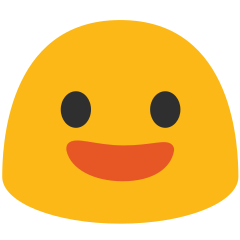} & 
    \includegraphics[width=0.1\textwidth]{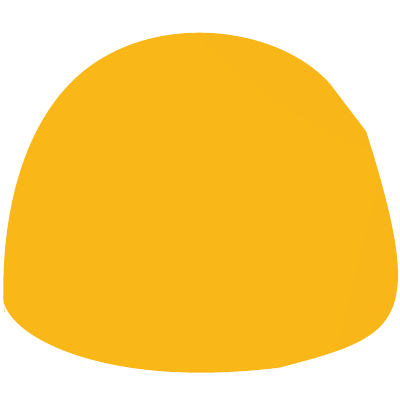} \\
    Input Image & DiffVG-Rand \\ 
    \includegraphics[width=0.1\textwidth]{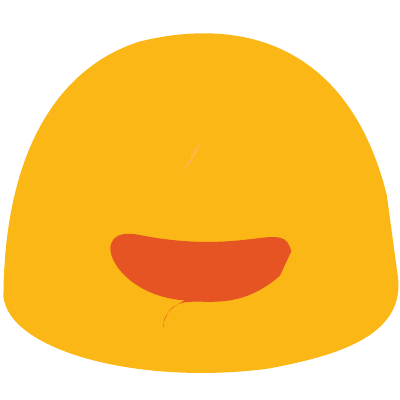} & 
    \includegraphics[width=0.1\textwidth] {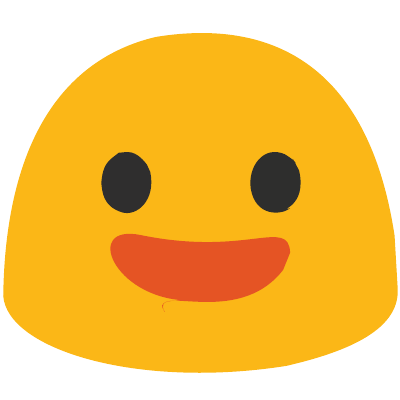} \\

DiffVG-DBSCAN & \textbf{Ours}  \\
    \end{tabular}
  \caption{\textbf{DBSCAN Initialization Contribution:} Results using 4 shapes. DiffVG is remarkably sensitive to the curve's initialization.}
  \label{fig:DBSCAN_contribution_emojis}
\end{figure}
\paragraph{\bf Image Editing}
Figure~\ref{fig:qualitative_editing_color} illustrates color editing for two example images. The left triplet displays (a) the original raster image, (b) the application of O\&R, and (c) a color edit in which the flame color has been changed. This color edit involved modifying \textit{a single shape} parameter. The right triplet demonstrates a color edit performed on the object's T-Shirt. In this case, multiple shapes were involved, but only the colors were changed. The underlying shape structure remained unaltered. 
As shown, our topology-aware method enables straightforward image editing.
\subsection{Implementation Details}\label{implementation}
We give a detailed description of the implementation details.
\paragraph{\textbf{{DBSCAN:}}} We use DBSCAN to cluster image colors. The DBSCAN parameters are fixed to $\epsilon=5$ gray levels, and $\text{min \textunderscore points}=20$. Since there is much redundancy in color space, it is enough to resize the image to $100 \times 100$ pixels and run DBSCAN on the decimated image. The cluster centers association is then projected back at the $240 \times 240$ image. An $\epsilon=5$ defines the distance in gray levels between points such that they are in the same neighborhood. $\text{min\_points}=20$ has implication on the smallest area for a connected component which is $20 \cdot \frac{240}{100} = 48$ pixels. 
\paragraph{\textbf{Optimization:}} We use the publicly available DiffVG implementation and its Pythonic interface. For qualitative evaluation, we follow~\cite{ma2022towards} and reshape images to $240 \times 240$. To optimize shape parameters, we use two instances of Adam~\cite{kingma2014adam} optimizers: one with a learning rate of $1.0$ for the \Besizer coordinates and the other, with a learning rate $10^{-2}$ to optimize for the RGBA parameters. Throughout the paper, we use $\mathcal{L}_{optimize} = \text{MSE}$ for quantitative evaluation. For qualitative evaluation we use the combination of $L_{1}$ and CLIP loss and our geometric loss: $\mathcal{L}_{optimize} = \alpha L_{1} + (1-\alpha)\mathcal{L}_{\text{CLIP}}$
\paragraph{\textbf{Reduce:}} Our Reduce steps follow an exponential decay schedule, that is, the number of shapes is halved in each iteration. We evaluate four optimization steps and three reduction steps. Midjourney target shapes are $32, 64, 128, 512$ and for the rest of the datasets, we follow LIVE~\cite{ma2022towards} for target shapes of $8, 16, 32 ~\text{and}~ 64$. 
\paragraph{\textbf{Geometric Loss:}} Our geometric loss has two hyper-parameters: $\lambda_{geometric}$ which weighs the contribution of the geometric loss for the optimization loss. Another is $\lambda_p$ which controls the balance between our intersubsections regularizer and the orientation regularizer.  In cases where the geometric loss is incorporated, we set its hyper-parameters to  $\lambda_{geometric}=0.01$ and $\lambda_{p}=10$.
\begin{figure}
\centering
\begin{tabular}{c}
    MSE ($\downarrow$ better) vs Runtime ($\leftarrow$ better) \\
     \includegraphics[width=0.45\textwidth]{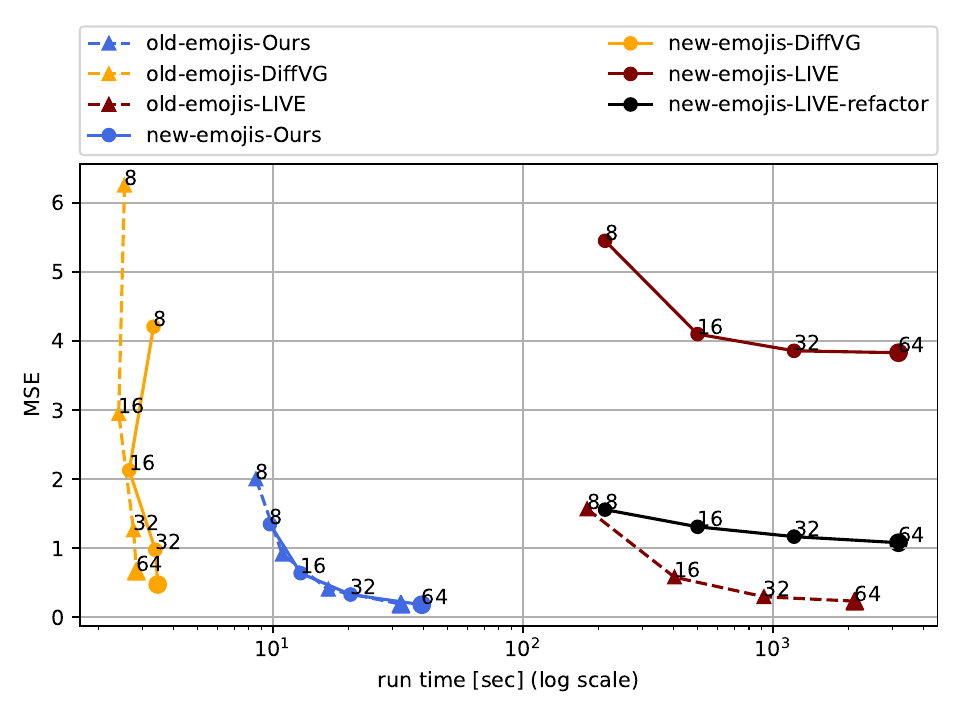} 
\end{tabular}
\caption{\textbf{MSE vs Runtime for the two Emojis datasets:} Colors code the different methods: DiffVG in yellow, LIVE in maroon and Ours in royal-blue. Triangles code the old emojis dataset and dots represent the new emojis dataset. The black line shows LIVE's result after we fixed their pre-processing bug. After the fix, we can see that our method reflects competitive results compared to LIVE for a low shapes budget. We can observe a MSE gap of $~3$ gray levels. }
\label{fig:live_with_runtime_with_bug}
\end{figure}
\paragraph{\textbf{Code:}} The evaluation of the optimization-based methods was conducted using their publicly available implementation.\footnote{\url{https://github.com/BachiLi/diffvg}}\footnote{\url{https://github.com/Picsart-AI-Research/LIVE-Layerwise-Image-Vectorization}}. We found a bug in LIVE's image loading process which degrades its performance when switching from the old emojis dataset to the new emojis dataset. When this bug was fixed, the MSE for 8 shapes was reduced from $5.3$ gray levels to $1.5$ gray levels. The bug-fix and Our methods implementation will be published. Figure~\ref{fig:live_with_runtime_with_bug} shows the MSE gap between the case with and without our bug-fix. We run our competitors using default parameters stated  by the authors.
\paragraph{\textbf{Benchmark Evaluation:}} Our goal for benchmark evaluation is to evaluate both the quality of reconstruction and the runtime of three methods: DiffVG, LIVE and Ours. For reconstruction error we use both pixel-wise (MSE) and Perceptual (LPIPS) losses for evaluation. As for runtime considerations, we run all three methods on an NVIDIA RTX A5000 GPU, i7-13700K CPU with Ubuntu 22.04.2 LTS.
We evaluate the runtime performance of DiffVG, LIVE, and our method. 
As DiffVG converges quickly to a pleasing solution, a fixed number of optimization iterations is unfair for runtime comparison. Thus, we introduce the following three conditions for the number of iterations: (1) an upper bound of $500$ iterations, (2) a lower bound of $50$ iterations, and 
(3) an early-stopping criteria, where we stop the optimization if the loss does not improve by $0.01\%$ from the previous iteration.
For our method, we allocate a total of $500$ iterations, spread across four steps: $150, 100, 100$, and $150$ iterations, respectively. We fix some bugs in the official LIVE implementation and modify it to support early stopping, for fair comparisons.
\paragraph{\textbf{Early Stopping:}}  We observed that DiffVG exhibits inconsistent convergence times across different images. The convergence time tends to increase with image complexity and the number of shapes involved. To address this issue, we propose an early stopping mechanism for DiffVG's optimization process. Specifically, we suggest a lower bound of $50$ iterations and an upper bound of $500$ iterations. Additionally, we monitor the loss function during optimization and halt the process if the loss fails to decrease by more than $10^{-2}\%$ compared to the previous iteration's loss value. To determine this threshold, we empirically analyzed the rate of loss decrease for multiple images and selected the value at which a non-monotonic decrease in loss improvement becomes apparent.
We have implemented early stopping for both DiffVG and Our method, as mentioned earlier. In the case of LIVE, we provide two evaluations: (1) LIVE with $500$ iterations per shape, and (2) LIVE with early stopping (LIVE-ES), utilizing a total of $500$ iterations. We have observed that LIVE terminates the optimization procedure once it reaches the minimum threshold of iteration steps. This behavior can be attributed to the slow convergence of LIVE's UDF loss towards the final shape. The gradual convergence leads to relatively small improvements in the loss from one iteration to the next, causing it to fall below the improvement condition for our early-stopping mechanism.
\subsection{Geometric Loss}
\begin{figure}
  \center
  \begin{tabular}{ccc}
    \begin{tabular}{c}
    \includegraphics[width=0.11\textwidth]{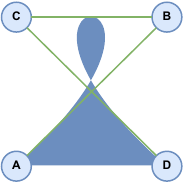}
    \end{tabular}
    & 
    \begin{tabular}{c}
    \includegraphics[width=0.11\textwidth]{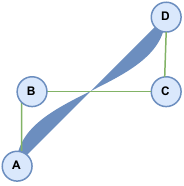}
    \end{tabular}
    & 
    \begin{tabular}{c}
    \includegraphics[width=0.11\textwidth]{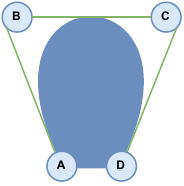}
    \end{tabular}
    \\
    (a) & (b) & (c)
    \end{tabular}  
  \caption{\textbf{Geometric Loss:} Our geometric constraint prevents curve intersubsections, resulting in the smoother topology of shapes. Our geometric loss term has three components, each one aiming to eliminate intersubsection cases: (b) Self-intersubsection occurs when AB intersects CD. In addition, Cross-intersubsections are prone to happen when: (c)  $[A, B, C]$ and $[B, C, D]$ have different orientations. (d) $\angle ABC$ or $\angle BCD$ are acute.}
  \label{fig:geometric_constraints_only}
\end{figure}
We refer the reader to Figure~\ref{fig:geometric_constraints_only}. Using the same mathematical formulation of cubic \Besizer curve (in order: A, B, C, and D) and using the orientation function ($O(A, B, C)$) as defined in the main paper, we can verify the same orientation for [A, B, C] and [B, C, D], or check whether AB intersects CD.
First, we define the soft Boolean functions AND and XOR:
\begin{equation}
AND(A, B) = A \cdot B
\end{equation}
\begin{equation}
XOR(A, B) = A + B - (A \cdot B)
\end{equation}
Using these functions, the same orientation for [A, B, C, D] can be defined as:
\begin{equation}
f_{orientation}(A, B, C, D) = AND(O(A, B, C), O(B, C, D))
\end{equation}
And the intersubsection of AB and CD occurs when:
\begin{equation}
\begin{split}
f_{intersect}(A, B, C, D) = \\ 
AND( &XOR(O(A, B, C), O(A, B, D)),  \\
     &XOR(O(C, D, A), O(C, D, B)))
\end{split}
\end{equation}
In cases where the geometric loss is incorporated, we set its hyperparameters to  $\lambda_{geometric}=0.01$ and $\lambda_{p}=10$.
\subsection{Quantitative Results: LPIPS}
\begin{figure*}
  \centering
  \begin{tabular}{ccccc}
  Old Emojis & New Emojis & Free-SVG & NFT-Apes dataset & Midjourney \\
    \includegraphics[width=0.18\textwidth]{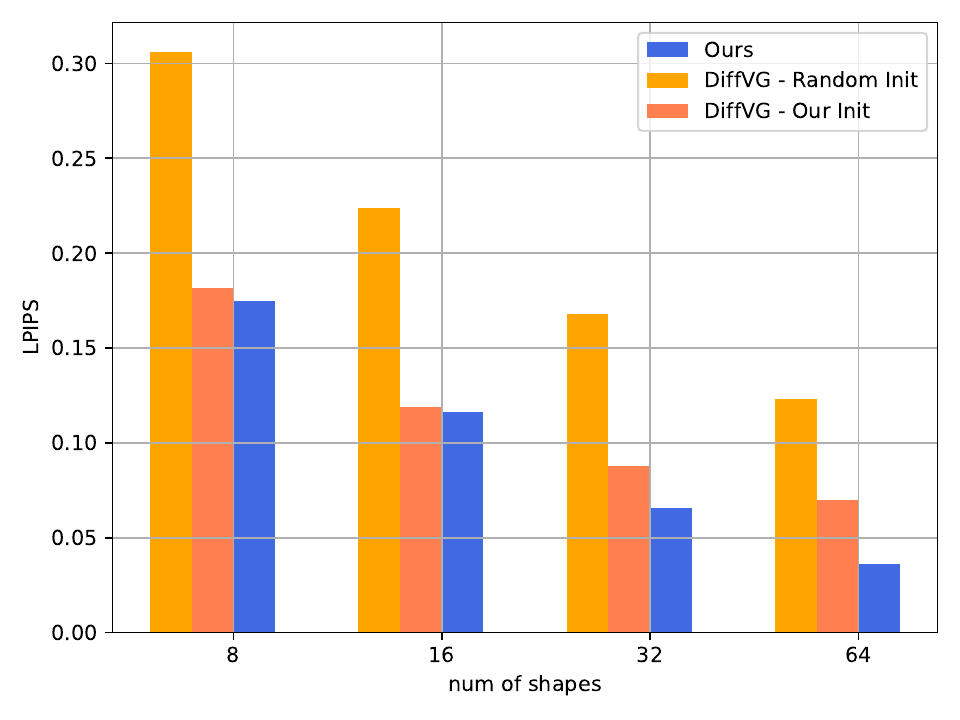} &
    \includegraphics[width=0.18\textwidth]{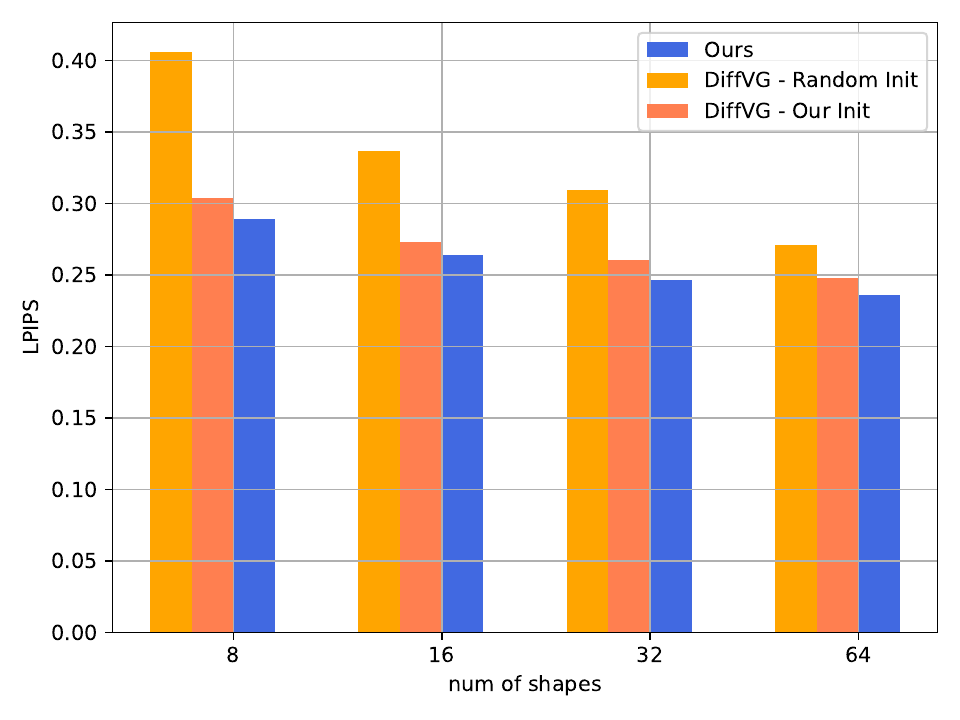} &
    \includegraphics[width=0.18\textwidth]{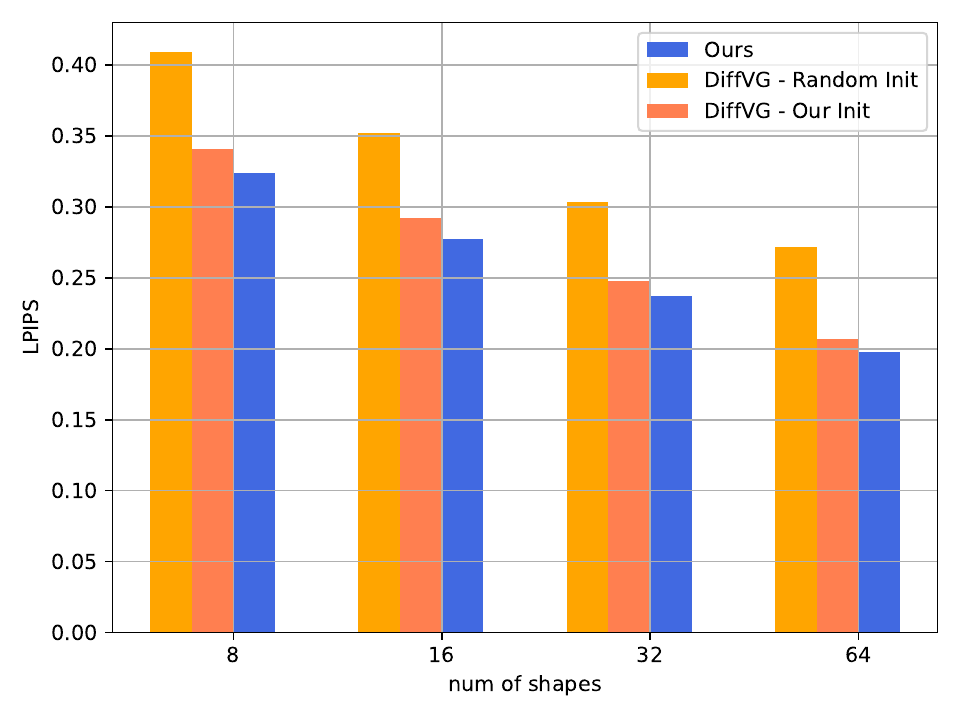} & 
    \includegraphics[width=0.18\textwidth]{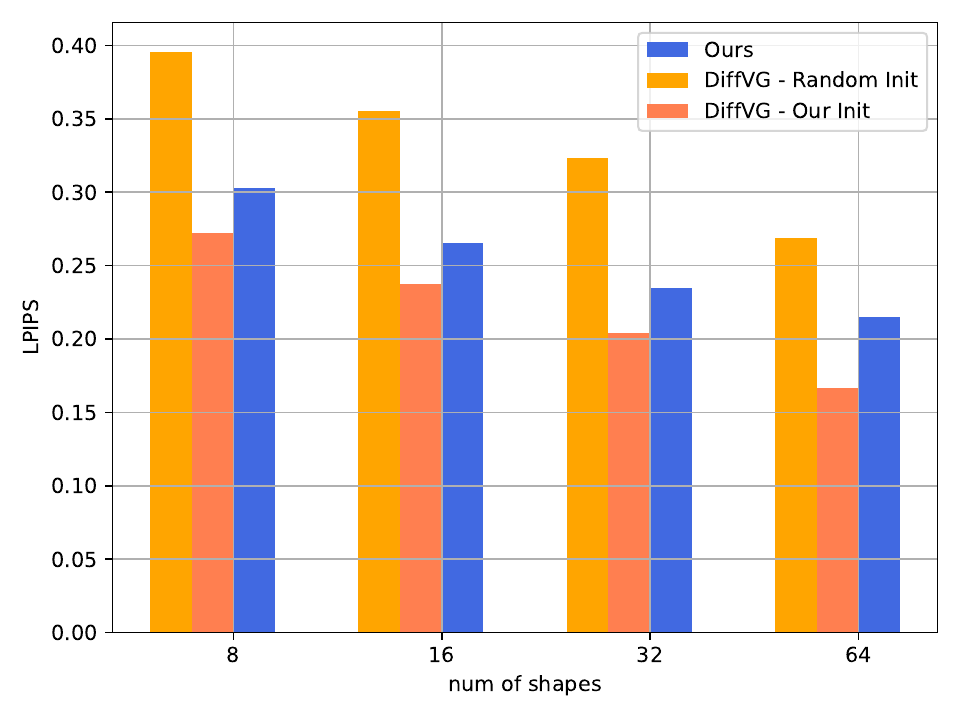} &
    \includegraphics[width=0.18\textwidth]{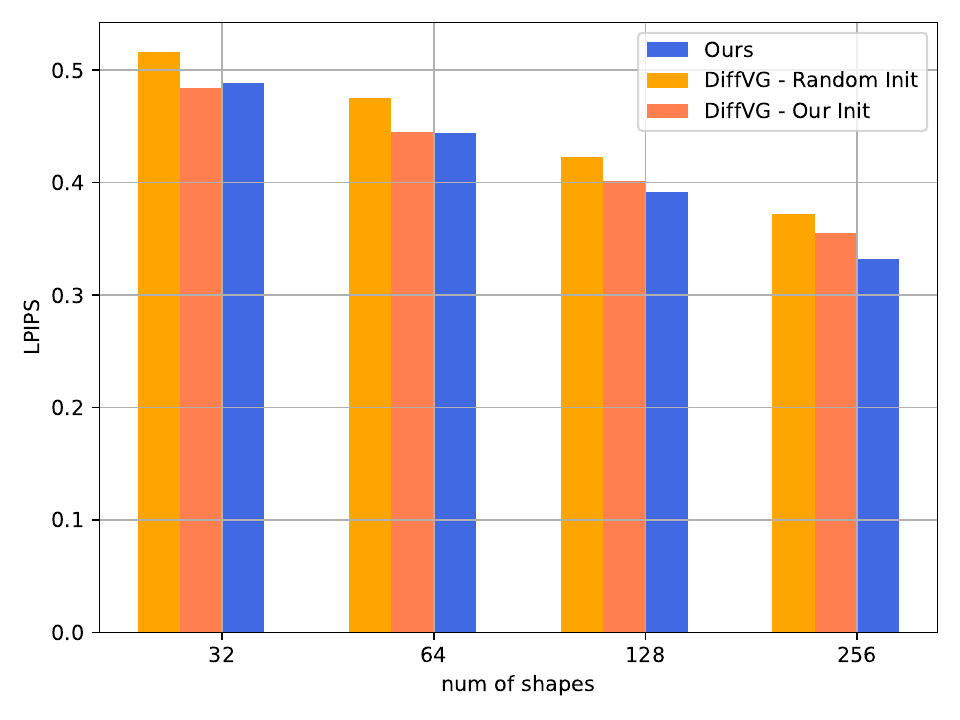}
    \end{tabular}
  \label{tab:my_table}
\caption{\textbf{LPIPS vs Number of shapes:}  Top row: MSE reconstruction error ($\downarrow$ better) vs the number of shapes in the vector representation. Bottom row: LPIPS score ($\downarrow$ better) vs the number of shapes. Our method consistently outperforms DiffVG for all datasets. Note that for the NFT-Apes dataset, our initialization imposes an inductive bias that slightly improves the performance of DiffVG when used in conjunction with our initialization.}
\label{fig:lpips_vs_num_shapes}
\end{figure*}
We evaluate the performance of our approach using two quantitative reconstruction metrics: (1) Mean Squared Error (MSE), as presented in the main paper, and (2) LPIPS (Learned Perceptual Image Patch Similarity)~\cite{zhang2018unreasonable}, a perceptual similarity metric that measures the similarity between two images. LPIPS leverages feature embeddings extracted from a VGG network [2] to compute the similarity between images. In Figure~\ref{fig:lpips_vs_num_shapes}, we depict the LPIPS score reconstruction error plotted against the number of shapes in the vector representation, with each method distinguished by a different color. Across all datasets, our method consistently surpasses DiffVG in terms of reconstruction quality when considering an equal number of shapes. 
\subsection{Further Results}
\begin{figure*}
\centering
\includegraphics[width=0.7\textwidth]{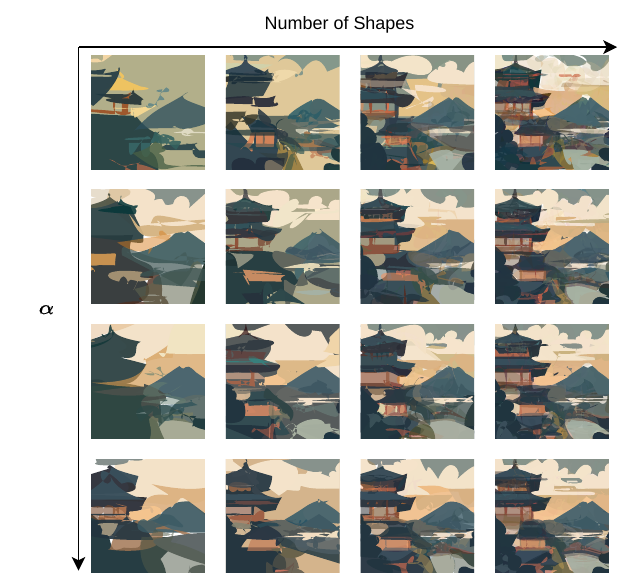}
\caption{\textbf{Abstraction Matrix:} Results obtained for a single input image using various values of $\alpha$ and number of shapes. $\alpha$ is the hyperparameter for $\mathcal{L}_{optimize} = \alpha L_{1} + (1-\alpha) \mathcal{L}_{\text{CLIP}}$.}
\label{fig:clip_matrix}
\end{figure*}

\begin{figure*}
    \centering
    \includegraphics[width=0.6\textwidth]{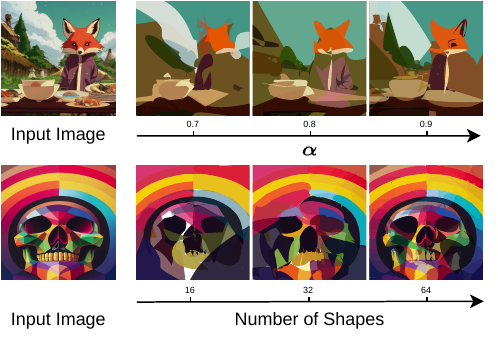}
    
    \caption{\textbf{Abstraction Examples:} Examples of different levels of abstraction using various $\alpha$ values and number of shapes. Lower $\alpha$ means more abstraction 
    }
    
    \label{fig:abstract_matrix_supp123}
\end{figure*}

\begin{figure*}
\centering
\includegraphics[width=0.8\textwidth]{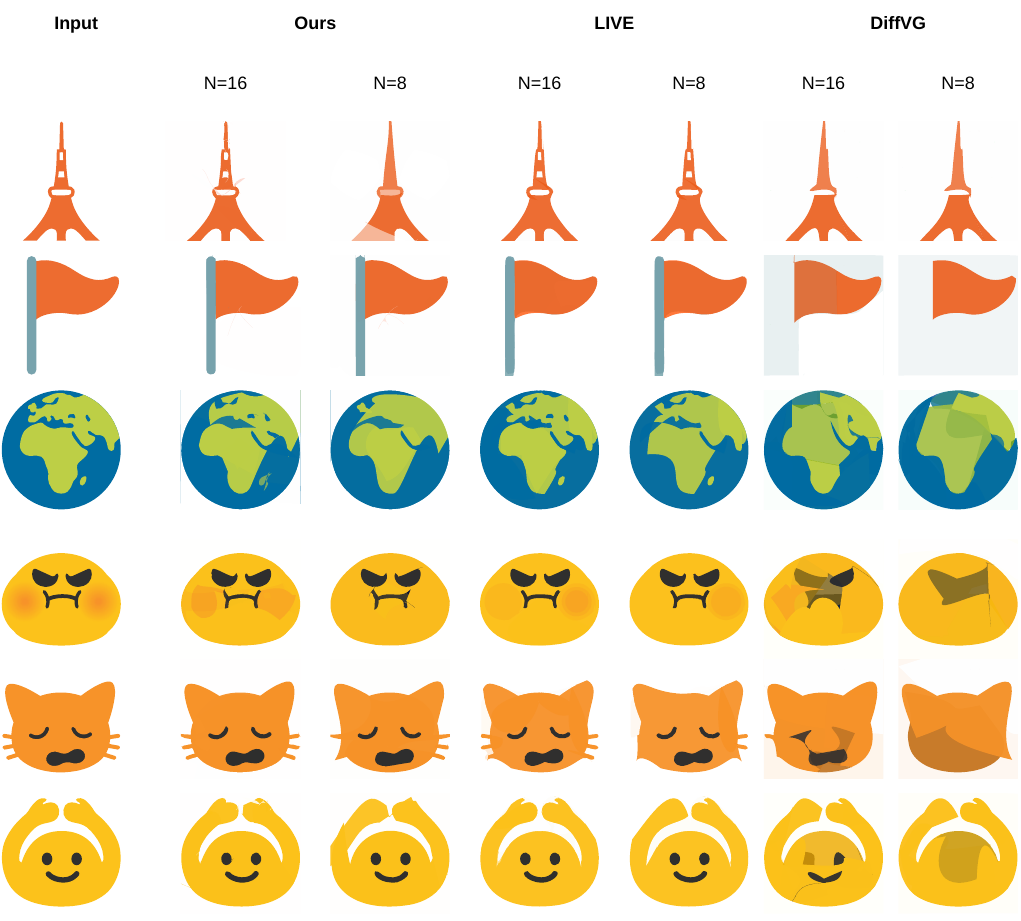}
\caption{\textbf{Qualitative Comparison:} We showcase the results of DiffVG, LIVE, and Our method on a collection of images from the former version of EMOJI dataset. Each image is compared using the same number of shapes.}
\label{fig:comparison_old_emoji}
\end{figure*}

\begin{figure*}[h]
\centering
\includegraphics[width=0.8\textwidth]{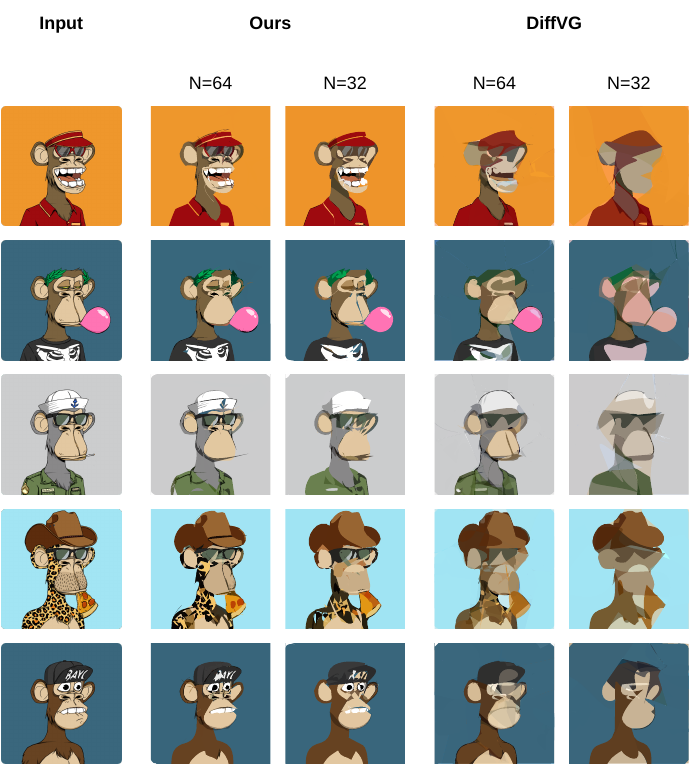}
\caption{\textbf{Qualitative Comparison:} We showcase the results of DiffVG and Our method on a collection of images from NFT-Apes dataset.}
\label{fig:comparison_apes_supp}
\end{figure*}

\begin{figure*}
\centering
\includegraphics[width=0.6\textwidth]{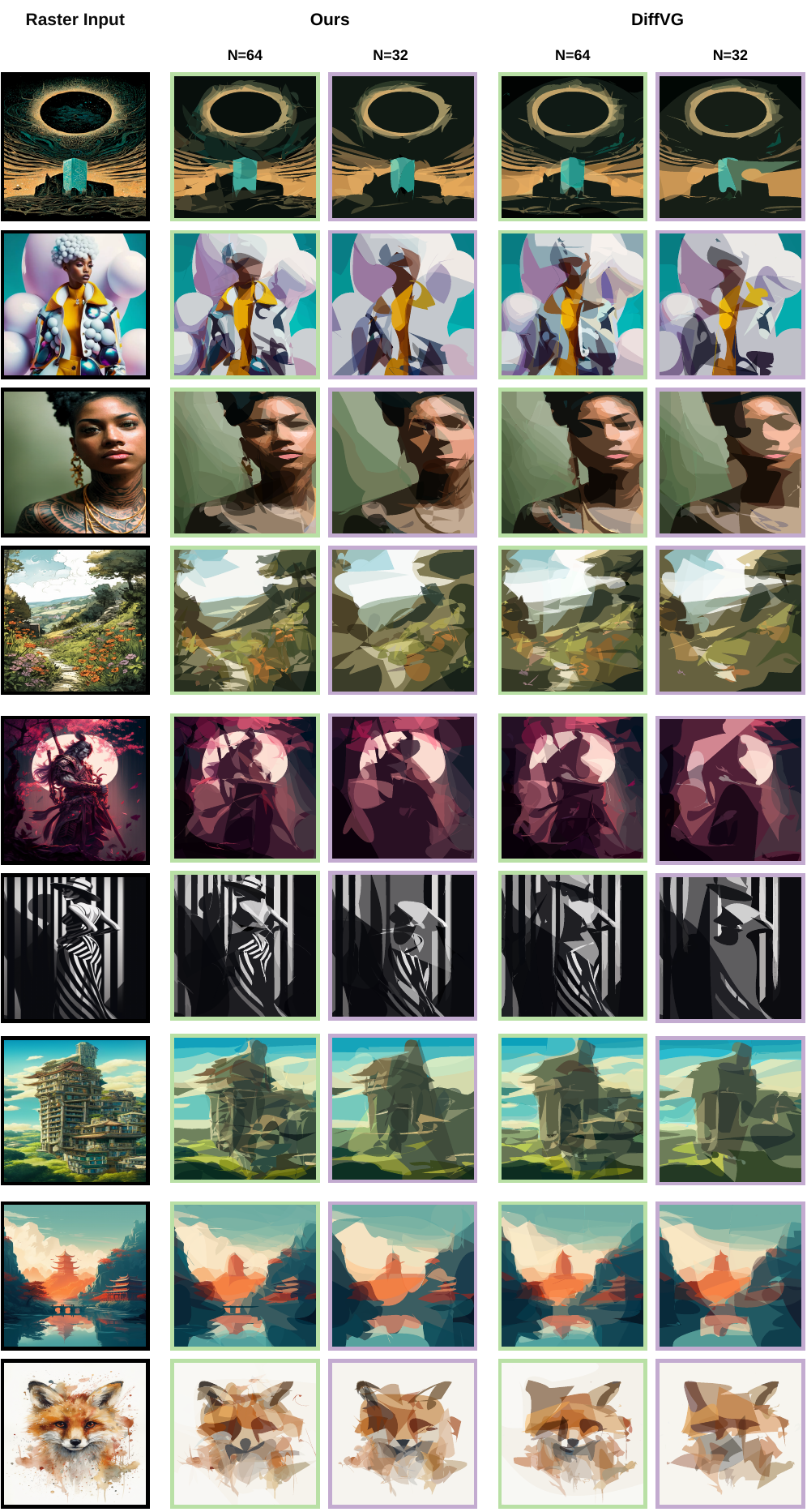}
\caption{\textbf{More Results:} We showcase more results of our method on a collection of images generated using Midjourney. Each image is vectorized using 64 32 shapes. (colored green and violet respectively)}
\label{fig:more_results_supp}
\end{figure*}

\begin{figure*}
  \centering
  \begin{tabular}{c}
     \includegraphics[width=0.5\textwidth]{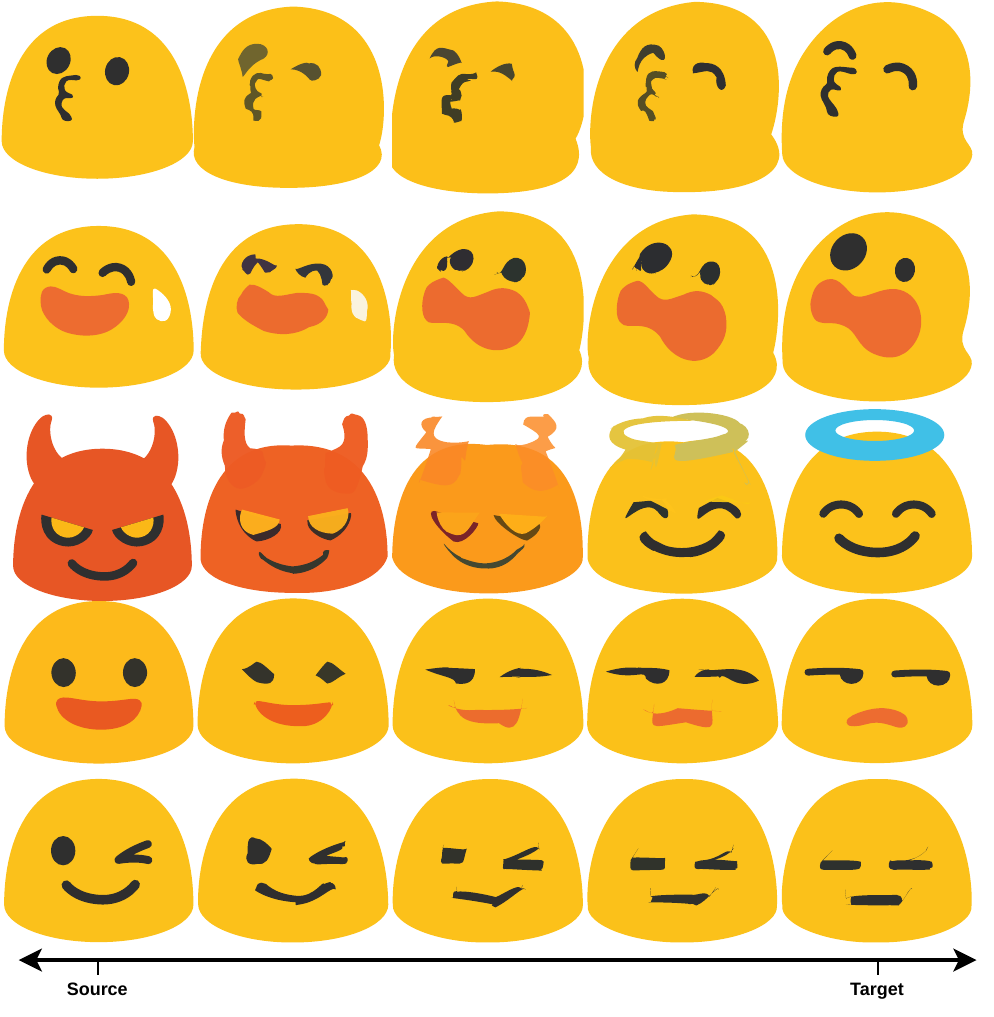} 
  \end{tabular}
  \caption{\textbf{Interpolation:} We perform an interpolation of emoji images from the rightmost column to the leftmost column. The initialization of the rightmost emoji is based on the vectorized result of the leftmost emoji. The intermediate images shown in between represent the optimization process's incremental results. These  constitute interpolated states between the source and the target images.. }
  \label{fig:interpolation_supp}
\end{figure*}

\paragraph{\textbf{Vectorization:}} Figure~\ref{fig:comparison_old_emoji} presents a qualitative comparison of DiffVG, LIVE, and our method for the vectorization task on the old emojis dataset. Figure~\ref{fig:comparison_apes_supp} showcases comparisons between DiffVG and Our method for vectorization on the NFT-Apes dataset. Additionally, Figure~\ref{fig:more_results_supp} displays vectorization results of images from the Midjourney dataset.
\paragraph{\textbf{Image Abstraction:}} In accordance with the approach presented in~\cite{vinker2022clipascene}, we showcase various levels of abstraction in our images. We achieve image abstraction by (1) varying the number of shapes in the vector representation and (2) adjusting a hyperparameter $\alpha$, which governs the balance between the $L_{1}$ loss and CLIP loss. We suggest using three CLIP loss terms: The first term describes the global image-to-image embedding loss - which reflects a contextualized term. The second term encourages low-level features extracted from the CLIP encoder to match. The third term encourages the prompt (text) embedding and the image embedding to match. $E$ is the image encoder, $E_{t}$ is the text encoder and $E^{(l)}$ is the attention map of the $\ell$-th layer.
\begin{equation}
    \begin{aligned}
    \mathcal{L}_{clip} & = \text{dist}\Big(E(I), E(\hat{I}) \Big)  \\
                        & + \lambda_{\text{CLIP-geo}} \cdot \sum_{\ell} \Big| {E}^{(\ell)}(I)-{E}^{(\ell)}(\hat{I}) \Big| _{2} \\
                        & + \lambda_{\text{CLIP-text}} \cdot  \text{dist}\Big(E_{t}(\text{text}), E(\hat{I})\Big)  \\
    \end{aligned}
    \label{eq:clip}
\end{equation}
We illustrate the individual effects of varying $\alpha$ and the number of shapes in Figure~\ref{fig:abstract_matrix_supp123}. Additionally, we showcase the combined effect of adjusting both parameters in Figure~\ref{fig:clip_matrix}. We kindly ask the reader to zoom in for details.
\paragraph{\textbf{Interpolation: }} We provide more examples of emojis interpolation in Figure~\ref{fig:interpolation_supp}. We ask the reader to zoom in for details.
\subsection{Datasets Showcase and Specifications}
\begin{table*}[h]
\tiny
\centering
\begin{tabular}{lllll}
32/emoji\_u1f609 & 512/emoji\_u1f60c & 32/emoji\_u1f637 & 72/emoji\_u1f619 & 512/emoji\_u1f620 \\ 
512/emoji\_u1f630 & 128/emoji\_u1f620 & 128/emoji\_u1f913 & 32/emoji\_u1f62f & 72/emoji\_u1f62b \\ 
32/emoji\_u1f629 & 32/emoji\_u1f627 & 32/emoji\_u1f62a & 32/emoji\_u1f603 & 72/emoji\_u1f603 \\ 
72/emoji\_u1f62a & 72/emoji\_u1f612 & 512/emoji\_u1f605 & 32/emoji\_u1f605 & 72/emoji\_u1f970 \\ 
72/emoji\_u1f623 & 512/emoji\_u1f61f & 512/emoji\_u1f606 & 128/emoji\_u1f62e & 128/emoji\_u1f627 \\ 
128/emoji\_u1f912 & 32/emoji\_u1f635 & 32/emoji\_u1f643 & 512/emoji\_u1f631 & 72/emoji\_u1f632 \\ 
128/emoji\_u1f61c & 128/emoji\_u1f612 & 128/emoji\_u1f60a & 512/emoji\_u1f912 & 72/emoji\_u1f610 \\ 
128/emoji\_u1f613 & 32/emoji\_u1f60f & 32/emoji\_u1f61f & 32/emoji\_u1f917 & 72/emoji\_u1f625 \\ 
72/emoji\_u1f616 & 72/emoji\_u1f975 & 128/emoji\_u1f925 & 128/emoji\_u1f917 & 512/emoji\_u1f611 \\ 
32/emoji\_u263a & 128/emoji\_u1f632 & 512/emoji\_u1f641 & 128/emoji\_u1f62b & 128/emoji\_u1f644 \\ 
512/emoji\_u1f61d & 512/emoji\_u1f642 & 512/emoji\_u1f602 & 32/emoji\_u1f621 & 512/emoji\_u1f615 \\ 
128/emoji\_u1f630 & 128/emoji\_u1f607 & 32/emoji\_u1f620 & 32/emoji\_u1f607 & 32/emoji\_u1f613 \\ 
72/emoji\_u1f60b & 128/emoji\_u1f636\_200d\_1f32b & 32/emoji\_u1f611 & 128/emoji\_u1f61f & 32/emoji\_u1f911 \\ 
32/emoji\_u1f618 & 32/emoji\_u1f62e & 32/emoji\_u1f600 & 128/emoji\_u1f606 & 32/emoji\_u1f622 \\ 
128/emoji\_u1f634 & 72/emoji\_u1f62f & 72/emoji\_u1f617 & 512/emoji\_u1fae2 & 128/emoji\_u1f61a \\ 
72/emoji\_u1f62c & 72/emoji\_u1f609 & 72/emoji\_u1f605 & 72/emoji\_u1f971 & 512/emoji\_u1f637 \\ 
512/emoji\_u1f913 & 128/emoji\_u1f619 & 32/emoji\_u1f633 & 32/emoji\_u1fae8 & 512/emoji\_u1f914 \\ 
128/emoji\_u1f636 & 72/emoji\_u1f60f & 72/emoji\_u1f600 & 72/emoji\_u1f635\_200d\_1f4ab & 32/emoji\_u1f913 \\ 
72/emoji\_u1f642 & 512/emoji\_u1f973 & 128/emoji\_u1f605 & 72/emoji\_u1f613 & 512/emoji\_u1f970 \\ 
128/emoji\_u1f629 & 72/emoji\_u1f620 & 72/emoji\_u1f611 & 512/emoji\_u1f627 & 32/emoji\_u1f636\_200d\_1f32b \\ 
32/emoji\_u1f631 & 32/emoji\_u1fae0 & 512/emoji\_u1f618 & 512/emoji\_u1f62d & 512/emoji\_u1f634 \\ 
512/emoji\_u1f636 & 512/emoji\_u1f62f & 32/emoji\_u1f610 & 72/emoji\_u1fae0 & 512/emoji\_u1f62b \\ 
512/emoji\_u1f910 & 72/emoji\_u1f60c & 512/emoji\_u1f600 & 32/emoji\_u1f62c & 512/emoji\_u1f60d \\ 
512/emoji\_u1f628 & 32/emoji\_u1f614 & 128/emoji\_u1f910 & 32/emoji\_u1f628 & 32/emoji\_u1f616 \\ 
32/emoji\_u1f60b & 128/emoji\_u1f626 & 72/emoji\_u1f915 & 512/emoji\_u1f915 & 512/emoji\_u1f925 \\ 
128/emoji\_u1f642 & 72/emoji\_u1f634 & 72/emoji\_u1f644 & 72/emoji\_u1f636 & 128/emoji\_u1f974 \\ 
72/emoji\_u1f622 & 128/emoji\_u1f621 & 128/emoji\_u1f975 & 32/emoji\_u1f972 & 72/emoji\_u1f637 \\ 
32/emoji\_u1f61a & 512/emoji\_u1fae0 & 72/emoji\_u1f629 & 32/emoji\_u1f974 & 512/emoji\_u1f632 \\ 
512/emoji\_u1f644 & 128/emoji\_u1f631 & 512/emoji\_u1f61c & 512/emoji\_u1f60e & 512/emoji\_u1f60f \\ 
72/emoji\_u1f633 & 512/emoji\_u1f923 & 128/emoji\_u1f633 & 128/emoji\_u1f617 & 72/emoji\_u1f604 \\ 
\hline\hline
emoji\_u1f6af & emoji\_u1f3a7 & emoji\_u1f5ff & emoji\_u1f631 & emoji\_u1f6a4 \\ 
emoji\_u1f6ae & emoji\_u1f627 & emoji\_u1f6a2 & emoji\_u1f478 & emoji\_u1f474 \\ 
emoji\_u1f618 & emoji\_u1f637 & emoji\_u1f6b0 & emoji\_u1f1ec\_1f1e7 & emoji\_u1f30e \\ 
emoji\_u1f4fc & emoji\_u1f622 & emoji\_u1f31e & emoji\_u1f3a8 & emoji\_u1f6a8 \\ 
emoji\_u1f602 & emoji\_u1f1ea\_1f1f8 & emoji\_u1f473 & emoji\_u1f612 & emoji\_u1f6a3 \\ 
emoji\_u1f605 & emoji\_u1f6a0 & emoji\_u1f6b2 & emoji\_u1f4fb & emoji\_u1f6a6 \\ 
emoji\_u1f639 & emoji\_u1f6b1 & emoji\_u1f635 & emoji\_u1f472 & emoji\_u1f3a4 \\ 
emoji\_u1f640 & emoji\_u1f33c & emoji\_u1f603 & emoji\_u1f479 & emoji\_u1f611 \\ 
emoji\_u1f33a & emoji\_u1f6ab & emoji\_u1f31b & emoji\_u1f620 & emoji\_u1f604 \\ 
emoji\_u1f1ee\_1f1f9 & emoji\_u1f609 & emoji\_u1f31f & emoji\_u1f636 & emoji\_u1f632 \\ 
emoji\_u1f614 & emoji\_u1f621 & emoji\_u1f5fe & emoji\_u1f638 & emoji\_u1f476 \\ 
emoji\_u1f467 & emoji\_u1f30f & emoji\_u1f628 & emoji\_u1f33d & emoji\_u1f31a \\ 
emoji\_u1f634 & emoji\_u1f6aa & emoji\_u1f633 & emoji\_u1f648 & emoji\_u1f3a1 \\ 
emoji\_u1f6a9 & emoji\_u1f33e & emoji\_u1f624 & emoji\_u1f645 & emoji\_u1f626 \\ 
emoji\_u1f475 & emoji\_u1f3a3 & emoji\_u1f470 & emoji\_u1f623 & emoji\_u1f6ac \\ 
emoji\_u1f30c & emoji\_u1f601 & emoji\_u1f468 & emoji\_u1f617 & emoji\_u1f3a5 \\ 
emoji\_u1f1ef\_1f1f5 & emoji\_u1f33f & emoji\_u1f469 & emoji\_u1f608 & emoji\_u1f471 \\ 
emoji\_u1f616 & emoji\_u1f649 & emoji\_u1f3a2 & emoji\_u1f5fd & emoji\_u1f629 \\ 
emoji\_u1f646 & emoji\_u1f477 & emoji\_u1f6a7 & emoji\_u1f1eb\_1f1f7 & emoji\_u1f466 \\ 
emoji\_u1f31d & emoji\_u1f34b & emoji\_u1f30d & emoji\_u1f610 & emoji\_u1f34a \\ 
emoji\_u1f31c & emoji\_u1f5fb & emoji\_u1f6a1 & emoji\_u1f606 & emoji\_u1f6ad \\ 
emoji\_u1f647 & emoji\_u1f30b & emoji\_u1f5fc & emoji\_u1f6b3 & emoji\_u1f625 \\ 
emoji\_u1f615 & emoji\_u1f600 & emoji\_u1f6a5 & emoji\_u1f3a0 & emoji\_u1f33b \\ 
emoji\_u1f619 & emoji\_u1f613 & emoji\_u1f3a6 & emoji\_u1f630 & emoji\_u1f607 \\ 
&&\\
\end{tabular}
\caption{\textbf{Emoji Dataset:} File names from noto-emoji dataset used for evaluation: (Top) new format, (Bottom) old format. The format of the new emojis dataset is resolution/emoji\_filename. }
\label{tab:emojis_names}
\end{table*}
Table~\ref{tab:emojis_names} explicitly report the images names from the two versions of Noto-Emojis dataset that we collect. The new emojis dataset supports different resolutions: $32\times32$, $72\times72$, $128\times128$, $512\times512$. We randomly selected images from all resolutions. The top subsection of Table~\ref{tab:emojis_names} shows the emojis names and their resolutions in a \textit{resolution/emojis\_filename} format. The bottom subsection of Table~\ref{tab:emojis_names} lists the emojis names collected from the old version of Noto-Emojis.
\begin{figure*}
\centering
\includegraphics[width=0.8\textwidth]{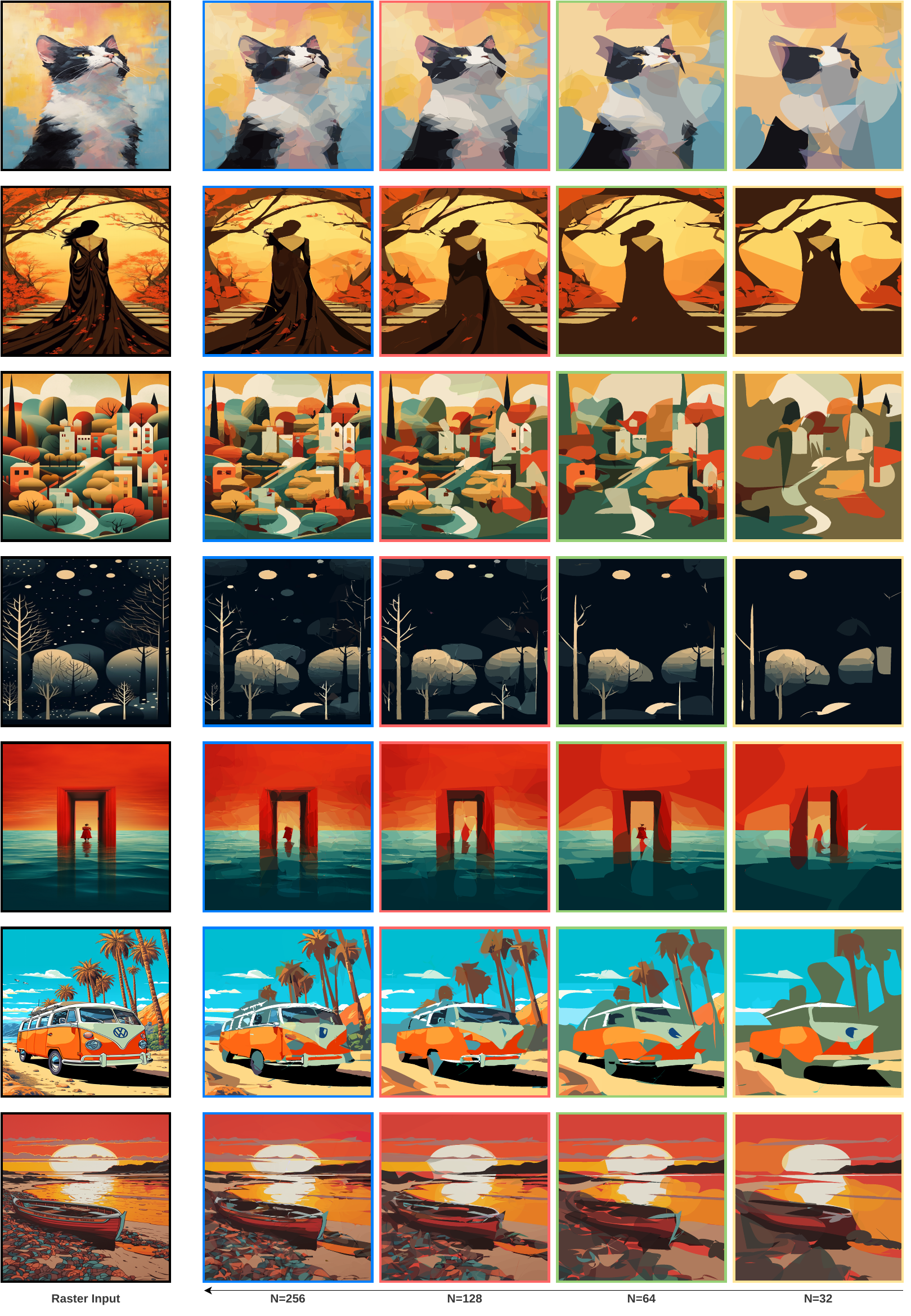}
\caption{\textbf{Additional Results:} We showcase more results of our method on a collection of images generated using Midjourney. Each image is vectorized using 256, 128, 64, and 32 shapes (shown in blue, red, green, and yellow, respectively).}
\label{fig:more_results}
\end{figure*}
\end{document}